\documentclass[lettersize,journal]{IEEEtran}

\usepackage{amsmath,amsfonts,amssymb}
\usepackage{algorithmic}
\usepackage{algorithm}
\usepackage{array}
\usepackage[caption=false,font=normalsize,labelfont=sf,textfont=sf]{subfig}
\usepackage{textcomp}
\usepackage{stfloats}
\usepackage{url}
\usepackage{verbatim}
\usepackage{graphicx}
\usepackage{cite}
\usepackage{booktabs}
\usepackage{multirow}
\usepackage{makecell}
\usepackage[hidelinks]{hyperref}

\begin{document}

\title{Rethinking Conditional Generation for Underwater Salient Object Detection}

\author{Hua Li, Yongjie Weng, Yutong Li, Zhiyuan Li, Runmin Cong,~\IEEEmembership{Senior Member,~IEEE}, and Sam Kwong,~\IEEEmembership{Fellow,~IEEE}%
\thanks{(Corresponding author: Runmin Cong.)}%
\thanks{Hua Li, Yongjie Weng, Yutong Li, and Zhiyuan Li are with the School of Computer Science and Technology, Hainan University, Haikou 570228, China 
(e-mail: lihua@hainanu.edu.cn; wengyongjie@hainanu.edu.cn; liyutong@hainanu.edu.cn; lizhiyuan@hainanu.edu.cn).}%
\thanks{Runmin Cong is with the School of Control Science and Engineering and the Key Laboratory of Machine Intelligence and System Control, Ministry of Education, Shandong University, Jinan, Shandong 250061, China 
(e-mail: rmcong@sdu.edu.cn).}%
\thanks{Sam Kwong is with the School of Data Science, Lingnan University, Hong Kong, China 
(e-mail: samkwong@ln.edu.hk).}%
}

\markboth{IEEE Transactions on Circuits and Systems for Video Technology}
{Li \MakeLowercase{\textit{et al.}}: Rethinking Conditional Generation for Underwater Salient Object Detection}

\maketitle

\begin{abstract}
Salient Object Detection in underwater images remains challenging due to low contrast, uneven illumination, and color distortion caused by scattering and absorption effects, which limit the effectiveness of conventional SOD methods in underwater environments. To address these challenges, we propose a \textbf{D}egradation-aware \textbf{C}onditional \textbf{G}eneration Network (DCGNet), specifically designed to construct reliable conditional features for underwater saliency generation. First, we design a Dynamic Multi-Granularity module (DMG) grounded in the human visual system to robustly detect salient objects of varying scales with blurred boundaries. Then, we develop an Underwater Physics-Prior module (UPP), which utilizes pseudo-depth guidance to estimate underwater light attenuation and backscatter, thereby restoring degradation-aware RGB features and mitigating color distortion and boundary ambiguity. Based on the physics-guided representation, we introduce an Underwater Spatial Gaussian module (USG), which constructs a spatial Gaussian saliency prior from the strongest guided response to enhance object-centered salient regions and suppress cluttered underwater backgrounds. In addition, a lightweight timestep-adaptive Diffusion Transformer (DiT) bottleneck is inserted into the denoising decoder to refine fused features at different diffusion timesteps. Comprehensive experiments on USOD10K, USOD, CSOD10K, MAS3K, and RMAS demonstrate that DCGNet significantly outperforms existing state-of-the-art methods, verifying its potential for complex underwater visual applications.

\end{abstract}


\section{Introduction}
\label{sec:intro}

\IEEEPARstart{U}{nderwater} Salient Object Detection (USOD) aims to locate and segment the most visually attractive objects from complex underwater scenes. As a fundamental task for underwater scene understanding, USOD supports many ocean-related applications, including underwater robot navigation, ecological monitoring, marine resource exploration, and infrastructure inspection \cite{jia2026vit,jin2024underwater}. Despite the progress of salient object detection in natural scenes, directly extending these methods to underwater environments remains challenging.

The difficulty mainly comes from the unique degradation mechanism of underwater imaging. Due to wavelength-dependent absorption, scattering, color distortion, and non-uniform illumination, underwater images often suffer from low contrast, blurred boundaries, and degraded visibility \cite{akkaynak2019sea,liu2020realworld,zhang2023underwater}. These degradations weaken foreground-background contrast, blur object contours, and damage the essential cues required by downstream perception and saliency detection \cite{fu2023learning,jin2024underwater}. Therefore, USOD requires not only strong semantic representation, but also degradation-aware and structure-preserving feature modeling.

Existing USOD methods mainly improve saliency representation from two aspects. One line of work introduces auxiliary cues, such as depth maps, boundary priors, or RGB-D fusion, to enhance object localization and boundary perception \cite{hong2023usod10k,jin2024underwater}. Another line adopts more powerful architectures, such as CNN-Transformer hybrid networks, to improve long-range dependency modeling and multi-scale feature interaction \cite{yuan2025if}. These discriminative methods have significantly promoted the development of USOD. However, most of them still follow a direct regression paradigm, where the saliency map is predicted from degraded observations in a single forward process. Since auxiliary cues such as depth and boundary maps are also estimated from degraded underwater images, the resulting conditional features may inherit unreliable contrast, distorted colors, blurred edges, and noisy depth responses. As a result, these methods often struggle to recover complete and boundary-consistent salient regions under severe underwater degradation.
\begin{figure}[t]
  \centering
  \includegraphics[width=1.0\linewidth]{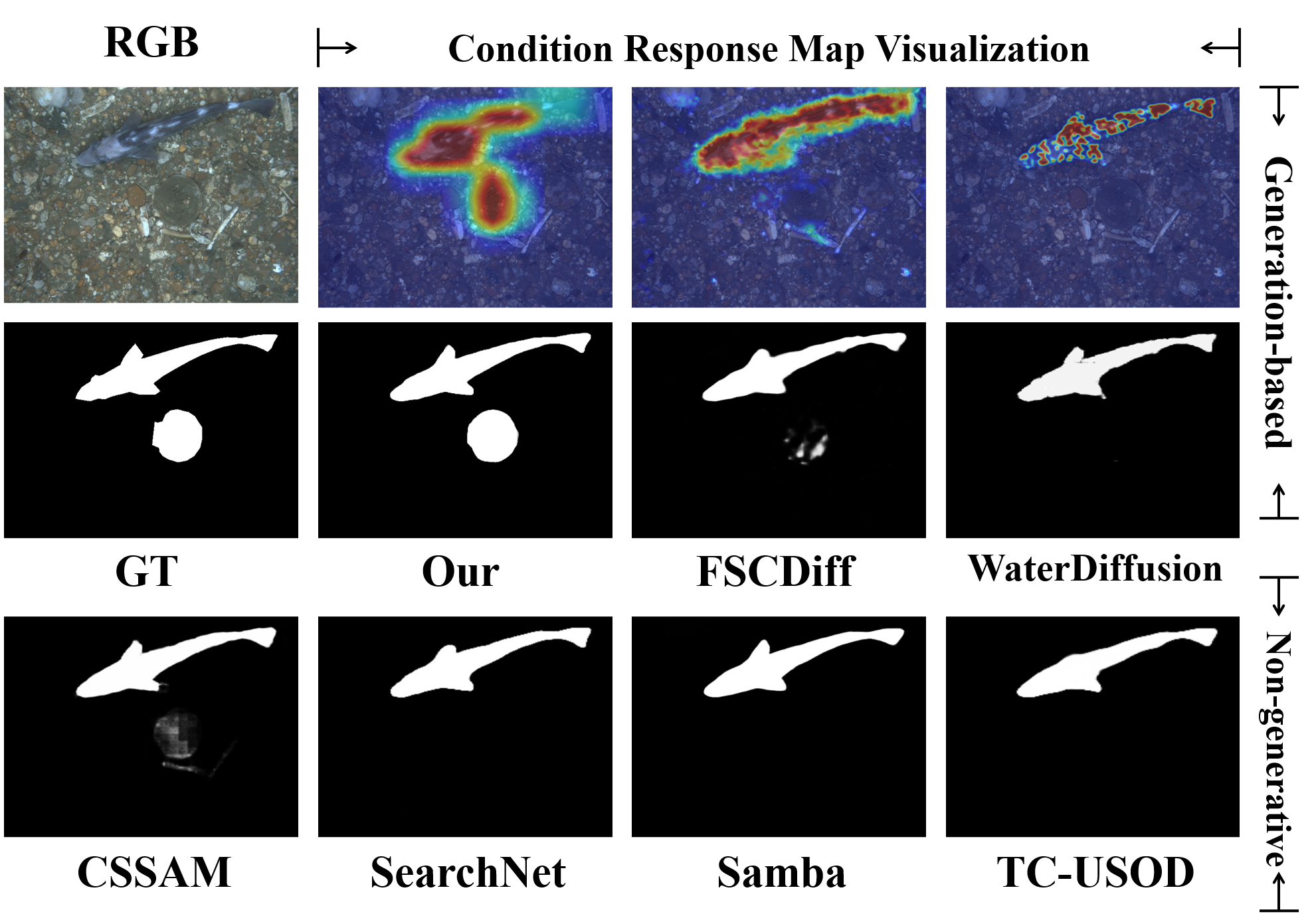}
  \caption{Visualization of conditional feature reliability in USOD. The first row shows conditional feature response maps of generative models, the second row shows their generated saliency masks, and the third row shows response maps of recent non-generative SOTA methods.}
  \label{fig:fig1}
\end{figure}
Recently, generative models have been introduced into USOD, shifting underwater saliency prediction from one-step mask regression to generation-driven mask modeling. Related studies also show the potential of generative modeling for reliable saliency prediction and underwater restoration \cite{mao2024generative,lu2024speedup}. WaterDiffusion \cite{chang2025waterdiffusion} formulates underwater saliency detection and image restoration within a prior-involved diffusion framework. However, its task setting differs from conventional saliency-only detection because the released protocol depends on additional restoration targets and medium-related priors. FSCDiff \cite{li2025fscdiff} explores frequency-spatial entangled conditional diffusion to alleviate insufficient representation and boundary deviation, while WaterFlow \cite{li2026waterflow} introduces underwater physics priors into a rectified flow framework. Different from direct regression methods, these generative approaches gradually generate saliency masks from noise through iterative denoising or flow under image-conditioned guidance. Therefore, conditional feature quality directly affects whether complete salient regions and accurate boundaries can be recovered.

Despite their promising performance, existing generative USOD methods reveal a more fundamental issue: the reliability of the conditional feature used to guide mask generation. In diffusion-based saliency prediction, the conditional feature is injected into multiple denoising steps rather than used only once. If the conditional feature is contaminated by color attenuation, backscattering, boundary blurring, or cluttered background responses, such errors may be repeatedly propagated, resulting in incomplete regions, unstable boundaries, or discontinuous structures. Fig.~\ref{fig:fig1} visualizes this issue. The first row shows the input underwater image and representative conditional feature response maps of generative models, while the second row presents their generated saliency masks. FSCDiff provides insufficient guidance for a shell-like salient object whose color is close to the surrounding background, leading to missed detection. WaterDiffusion produces scattered conditional feature responses, and the generated mask contains slight holes instead of a complete object region. The third row further shows response maps from recent non-generative SOTA methods, where missing regions or residual background activations can still appear under severe degradation. These observations suggest that the key problem is no longer merely how to generate a mask, but how to construct reliable, degradation-aware, and structure-preserving conditional features before generation. Accordingly, this work rethinks conditional generation from the perspective of conditional feature reliability.

To construct such conditional features, three challenges need to be addressed. First, multi-granularity structures are difficult to balance: underwater objects vary in scale, while their boundaries are often blurred by scattering. A single-granularity conditional representation is insufficient to capture both global context and local details. Second, spatial degradation is difficult to perceive because color attenuation and backscattering are spatially non-uniform, and contaminated regions may continuously mislead the generative process. Third, region completeness is difficult to preserve. Cluttered backgrounds can trigger local high responses, leading to fragmented or discontinuous saliency masks. Structural constraints are therefore required for complete object-centered predictions.

To address these challenges, we propose the Degradation-aware Conditional Generation Network (DCGNet) for conditional diffusion-based USOD. The core idea of DCGNet is to enhance the conditional features before mask generation, so that the denoising network can be guided by reliable underwater-aware features. Specifically, we first design a Dynamic Multi-Granularity (DMG) module to provide both coarse global localization cues and fine boundary details for diffusion guidance. Then, we introduce an Underwater Physics-Prior (UPP) module, which estimates degradation-aware features under pseudo-depth guidance and suppresses underwater backscatter and attenuation interference at the feature level. Based on the physics-guided representation, we further propose an Underwater Spatial Gaussian (USG) module, which constructs an object-centered soft spatial prior from the strongest guided response, encouraging region completeness and suppressing fragmented background activations. These modules improve diffusion guidance by providing more accurate localization, cleaner degradation-aware features, and stronger spatial continuity. In addition, we insert a lightweight timestep-adaptive DiT bottleneck into the denoising decoder to adjust global semantic reasoning and local boundary refinement at different diffusion stages.

The main contributions of this work are summarized as follows:

\begin{itemize}
\item We propose the Degradation-aware Conditional Generation Network (DCGNet) for USOD, which rethinks conditional generation from the perspective of conditional feature reliability and constructs degradation-aware guidance for saliency mask generation.

\item We design a Dynamic Multi-Granularity (DMG) module to capture global context and local boundary details, improving conditional representation under scale variations, weak contrast, and blurred underwater boundaries.

\item We introduce Underwater Physics-Prior (UPP) and Underwater Spatial Gaussian (USG) modules to model underwater degradation and object-centered spatial priors. In addition, a lightweight timestep-adaptive Diffusion Transformer (DiT) bottleneck is embedded into the denoising decoder for timestep-aware refinement.

\item Extensive experiments on USOD10K, USOD, CSOD10K, MAS3K, and RMAS demonstrate that DCGNet achieves state-of-the-art performance and exhibits strong robustness in complex underwater scenes.
\end{itemize}

\begin{figure*}[t]
    \centering
    \includegraphics[width=1.0\linewidth]{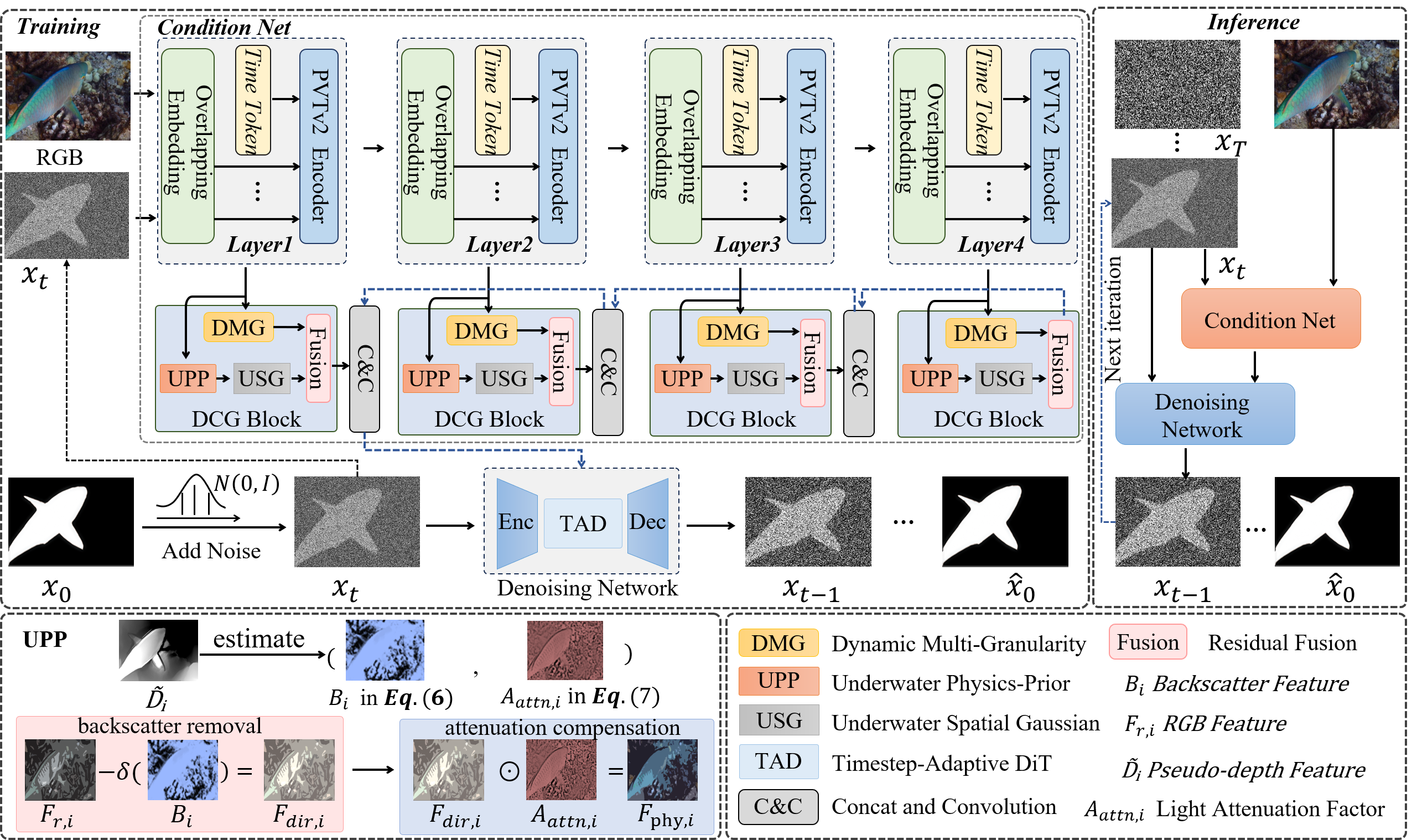}
    \caption{Overall architecture of DCGNet. The Condition Net extracts multi-level features through PVTv2 encoder stages and repeated DCG Blocks (Degradation-aware Condition Guidance), where DMG, UPP, and USG are fused to construct reliable degradation-aware conditional features. Given the noisy mask \(x_t\) and the extracted conditional features, the Denoising Network with a timestep-adaptive DiT bottleneck (TAD) progressively recovers the saliency mask \(\hat{x}_0\).}
    \label{fig:fig_overview}
    \end{figure*}
\section{Related Work}
\subsection{Underwater Salient Object Detection}
With the rapid development of deep learning, significant progress has been made in Salient Object Detection (SOD) in natural scenes. However, deep USOD is still in its infancy. Islam et al. \cite{islam2020svam} first proposed the USOD dataset, which contains 300 images and is limited in reflecting real-world underwater complexity. Hong et al. \cite{hong2023usod10k} proposed the first large-scale USOD dataset, USOD10K, and designed the benchmark method TC-USOD. Existing methods still face challenges under image degradation and complex underwater environments. Jin et al. \cite{jin2024underwater} introduced a dual-stage self-paced learning strategy and a depth emphasis module to refine depth guidance. Yuan et al. \cite{yuan2025if} proposed IF-USOD by combining CNN and Transformer architectures with interactive feature enhancement. Zhou et al. \cite{zhou2026turbidity} proposed a turbidity--similarity decoupling framework with feature-consistent mutual learning. Despite these advances, USOD remains challenging due to the unique characteristics of underwater environments.

\subsection{Diffusion Model}
Diffusion models have recently shown remarkable performance in various vision tasks, including image generation \cite{zhu2023conditional,mei2024codi}, deblurring \cite{feng2025residual,chen2023hierarchical}, image segmentation \cite{wu2024medsegdiff}, object detection \cite{chen2024camodiffusion,sun2025conditional}, and super-resolution reconstruction \cite{moser2024diffusion}. Different from deterministic prediction networks, diffusion models progressively recover clean targets from noisy samples through an iterative denoising process, providing a flexible probabilistic framework for dense prediction tasks.

Recent studies have further explored Transformer-based diffusion architectures. DiT \cite{peebles2023scalable} shows that Transformer blocks can serve as effective denoising backbones with strong global modeling and timestep-conditioned modulation, which is useful for dense prediction requiring both semantic reasoning and structural refinement. In underwater vision, WaterDiffusion \cite{chang2025waterdiffusion} introduces a prior-involved unrolling diffusion framework for joint underwater saliency detection and visual restoration, showing the potential of diffusion models in underwater perception. Nevertheless, because it jointly considers restoration and saliency detection and requires additional restoration/medium-related supervision, it is not a direct conventional RGB-to-mask saliency baseline. In contrast, our work focuses on saliency mask generation and strengthens the diffusion process by constructing more reliable degradation-aware and structure-preserving conditional features.

\begin{figure*}[t]
    \centering
    \includegraphics[width=0.9\linewidth]{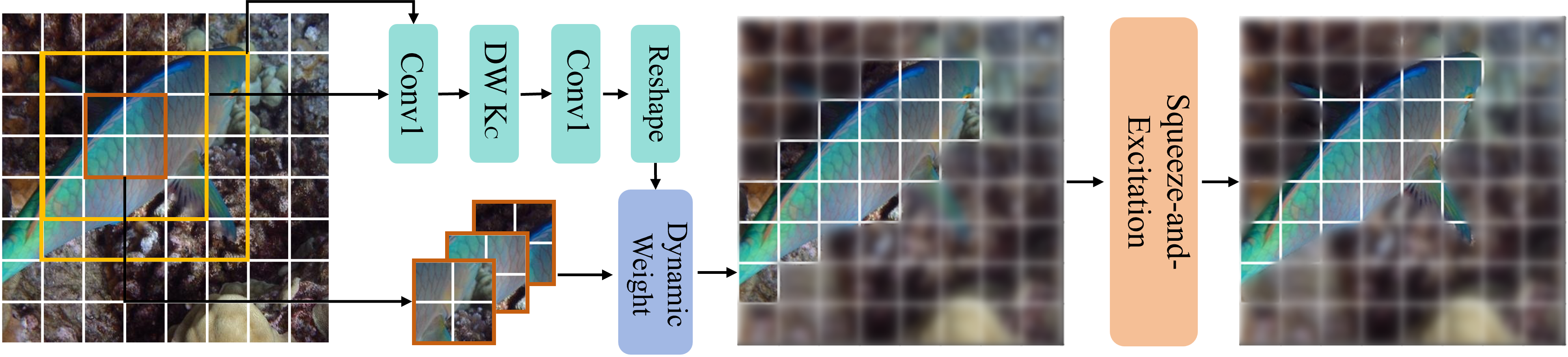}
    \caption{Overall architecture of Dynamic Multi-Granularity. By dynamically generating large-kernel weights, performing multi-scale feature aggregation, and integrating the squeeze-and-excitation channel attention mechanism.}
    \label{fig:fig3}
\end{figure*}

\section{Method}

\subsection{Overall Architecture}
As shown in Fig. \ref{fig:fig_overview}, the proposed DCGNet consists of two main components: a Condition Net for underwater feature extraction and a Denoising Network for mask generation. The architecture employs a hierarchical encoder based on PVTv2 \cite{wang2022pvt}.
The Condition Net comprises four hierarchical stages, each equipped with a DCG Block (Degradation-aware Condition Guidance) that integrates DMG, UPP, USG, and feature fusion to construct reliable degradation-aware conditional features. At the $i$-th stage ($i \in \{1, 2, 3, 4\}$), features $F_{i} \in \mathbb{R}^{C_i \times H_i \times W_i}$ are extracted. Three key modules process these features: (1) DMG uses multi-scale large-kernel perception and small-kernel dynamic aggregation to capture global context and local details, enhancing boundary extraction across multiple scales. (2) UPP estimates backscatter and light attenuation from pseudo-depth guidance, restoring degradation-aware structural and color cues for underwater features. (3) USG constructs a Gaussian saliency prior from the physics-guided representation through SpatialGSA, enhancing object-centered salient regions and reducing background interference. For diffusion-based mask generation, these modules provide complementary benefits: DMG stabilizes multi-scale object localization, UPP reduces degradation noise in the condition, and USG offers a soft spatial prior that encourages coherent denoising toward complete salient regions.
During training, Gaussian noise $\mathcal{N}(0, I)$ is progressively added to the ground-truth mask $x_0$ following the diffusion schedule, generating noisy samples $x_t$. The Denoising Network learns the reverse mapping from $x_t$ to $x_{t-1}$ via an encoder-decoder architecture, guided by conditional features from the Condition Net. During inference, starting from pure noise $x_T$, the network iteratively denoises to produce the final prediction $\hat{x}_0$.

In the Denoising Network, the fused conditional features are combined with the noisy mask feature $x_t$ to guide the reverse diffusion process. To better adapt the decoder to different denoising stages, we insert a lightweight Diffusion Transformer (DiT)-style bottleneck after feature-mask fusion. The timestep embedding is used to modulate both the self-attention and MLP branches, allowing the decoder to emphasize global semantic localization when the noise level is high and refine object boundaries when the noise level is low.

\subsection{Dynamic Multi-Granularity}

The root cause of saliency lies in the selective attention of the human visual system, which automatically prioritizes salient regions while ignoring redundant background information. However, underwater environments present unique challenges: light scattering and absorption degrade object saliency, causing blurred contours and reduced boundary contrast. This phenomenon hinders accurate foreground-background separation. To address this, we propose a Dynamic Multi-Granularity (DMG) module inspired by human visual principles \cite{yang2021focal,shi2024transnext,wang2025lsnet}. DMG employs multi-scale large kernels to perceive global context at coarse-grained scales while focusing on local details at fine-grained scales to enhance boundary extraction.
As shown in Fig. \ref{fig:fig3}, DMG generates spatially-varying large-kernel weights dynamically, aggregates local features via small-kernel unfold-fold operations, and recalibrates channel responses through squeeze-and-excitation, thereby enlarging the effective receptive field without heavy computational cost. For input feature $\mathbf{x} \in \mathbb{R}^{B \times C \times H \times W}$ (where $B$, $C$, $H$, $W$ denote batch size, channels, height, and width), DMG constructs parallel paths with kernel sizes $lks \in \{7, 11\}$, each performing channel reduction, depth-wise separable convolution, and dynamic weight generation.

The output weights from multi-scale paths are fused by averaging and normalized, which can be formulated as:
\begin{equation}
\mathbf{w} = \text{GroupNorm}\left( \frac{1}{N} \sum_{i=1}^N \mathcal{F}_{lks}^i(\mathbf{x}) \right),
\end{equation}
where $\mathcal{F}_{lks}^i(\cdot)$ denotes the feature transformation of the $i$-th path (including depth-wise separable convolution and channel adjustment), $N$ is the number of paths, and the fused weight is reshaped to $\mathbf{w} \in \mathbb{R}^{B \times g \times (C/g) \times s^2 \times H \times W}$ (with $g$ as the number of groups and $s$ as the small kernel size).

Subsequently, the previously generated dynamic weights are used to achieve adaptive feature aggregation. For the input feature $\mathbf{x}$, sliding window operations first extract small kernel neighborhood patches. The aggregation process is defined as:
\begin{equation}
\mathbf{y} = \text{Reshape}\left( \sum_{k=1}^{s^2} \text{patches}_k \cdot \mathbf{w}_k \right),
\end{equation}
where $\text{patches} = \text{unfold}(\mathbf{x}, k=s, p=(s-1)/2) \in \mathbb{R}^{B \times g \times (C/g) \times s^2 \times H \times W}$ are the neighborhood features extracted by sliding window. After aggregation and reshaping, the result becomes $\mathbf{y} \in \mathbb{R}^{B \times C \times H \times W}$, completing the dynamic convolution operation.

To enhance key channel features, the module incorporates the squeeze-and-excitation channel attention mechanism. Channel weights are learned through global average pooling and two-level fully connected layers:
\begin{equation}
\mathbf{a} = \sigma\left( \text{FC}_2\left( \text{ReLU}\left( \text{FC}_1\left( \text{AvgPool2d}(\mathbf{x}) \right) \right) \right) \right),
\end{equation}
where $\sigma$ is the Sigmoid function, $\text{FC}_1: C \to C/r$ and $\text{FC}_2: C/r \to C$ handle channel dimension scaling (with $r$ as the reduction ratio). The final output is the element-wise multiplication of the feature map and attention weights:
\begin{equation}
\mathbf{x}' = \mathbf{x} + \mathrm{BN}(\mathbf{y} \odot \mathbf{a}),
\end{equation}
where the residual connection stabilizes optimization and preserves the original representation while the dynamically aggregated feature $\mathbf{y}$ is recalibrated by channel attention.

\subsection{Underwater Physics-Prior}
\label{sec:upp}

Although DMG improves multi-scale feature representation, underwater-specific degradations such as wavelength-dependent attenuation and backscatter still weaken the discriminability of salient objects. These degradations usually cause color distortion, contrast reduction, and boundary blurring, making the subsequent spatial saliency enhancement less reliable. Therefore, before constructing the spatial Gaussian prior, we introduce the Underwater Physics-Prior module (UPP) to recover degradation-aware structural and color cues. UPP employs pseudo-depth guidance and learnable feature-level physical approximation to model underwater backscatter removal and attenuation compensation.

Sea-Thru \cite{akkaynak2019sea} models the observed underwater image $I_c$ in color channel $c \in \{R, G, B\}$ as a combination of direct transmission and backscatter:
\begin{equation}
I_c = J_c \cdot e^{-\beta_A^c \cdot z} + B_c^\infty \cdot \left(1 - e^{-\beta_B^c \cdot z}\right),
\label{eq:seathru_model}
\end{equation}
where $z$ is the range between the camera and objects along the line of sight. $B_c^\infty$ represents the veiling light, and $J_c$ denotes the unattenuated scene radiance at the camera location. The coefficients $\beta_A^c$ and $\beta_B^c$ depend on camera system and environmental parameters, such as range, reflectance, ambient illumination, camera spectral response, and water optical properties.

In our network, the exact physical depth is not available. Therefore, given the stage-wise RGB feature $F_i \in \mathbb{R}^{C_i \times H_i \times W_i}$, we first estimate a lightweight pseudo-depth map from the RGB feature to provide structure-aware guidance for underwater physics-prior modeling:
\begin{equation}
\begin{gathered}
\tilde{D}_i =
\sigma(\phi_{1\times1}^{d}(F_i)) \odot
\sigma\!\left(\phi_{5\times5}^{d}\left(\phi_{1\times1}^{d}(F_i)\right)\right),\\[3pt]
F_{d,i} = \mathrm{Expand}(\tilde{D}_i),
\end{gathered}
\label{eq:pseudo_depth}
\end{equation}
where $\tilde{D}_i \in \mathbb{R}^{1 \times H_i \times W_i}$ denotes the estimated single-channel pseudo-depth map, $F_{d,i} \in \mathbb{R}^{C_i \times H_i \times W_i}$ denotes the expanded pseudo-depth feature, $\mathrm{Expand}(\cdot)$ repeats the single-channel pseudo-depth map along the channel dimension, and $\sigma(\cdot)$ denotes the sigmoid function. This pseudo-depth estimation is a lightweight auxiliary operation used to provide degradation-aware structural guidance rather than an independent depth prediction task.

Given the RGB feature $F_{r,i}$ and the expanded pseudo-depth feature $F_{d,i}$, UPP estimates the spatially varying backscatter and obtains the direct component as follows:
\begin{equation}
\begin{gathered}
\beta_{\mathrm{back},i}=\mathrm{ReLU}\!\bigl(\phi_{\mathrm{back}}(F_{d,i})\bigr),\\[3pt]
B_i=B_\infty\!\bigl(1-\exp(-\beta_{\mathrm{back},i})\bigr),\\[3pt]
F_{\mathrm{dir}, i} = F_{r,i} - \delta\left(B_i\right),
\end{gathered}
\label{eq:backscatter_estimation}
\end{equation}
where $\phi_{\mathrm{back}}(\cdot)$ denotes a $1\times1$ convolution, $B_\infty \in \mathbb{R}^{C_i \times 1 \times 1}$ is a learnable ambient light parameter, and $\delta(\cdot)$ denotes the sigmoid function used to bound the estimated backscatter. This operation suppresses the veiling-light component caused by underwater scattering and obtains a feature-level approximation of the direct transmission component.

To compensate for wavelength-dependent absorption, we estimate a spatially varying attenuation coefficient from the pseudo-depth feature and obtain the physics-enhanced feature:
\begin{equation}
\begin{gathered}
\beta_{\mathrm{attn}, i} = \exp\left(-\mathrm{ReLU}\left(\phi_{\mathrm{attn}}(F_{d,i})\right)\right),\\[3pt]
A_{\mathrm{attn}, i} = \exp\left(
\mathrm{clip}
\left(
\mathrm{ReLU}\left(F_{d,i} \odot \beta_{\mathrm{attn},i} \odot w_a\right), 0, \tau
\right)
\right),\\[3pt]
F_{\mathrm{phy},i} = F_{\mathrm{dir}, i} \odot A_{\mathrm{attn}, i} \cdot w_s,
\end{gathered}
\label{eq:attenuation_and_enhancement}
\end{equation}
where $\phi_{\mathrm{attn}}(\cdot)$ denotes a $1\times1$ convolution, $w_a \in \mathbb{R}^{C_i \times 1 \times 1}$ controls the strength of attenuation compensation, $\tau$ is an upper bound used to avoid unstable exponential amplification, and $w_s \in \mathbb{R}^{1 \times 1 \times 1}$ controls the overall strength of illumination compensation. The output $F_{\mathrm{phy},i}$ is used as a physics-guided representation for subsequent spatial Gaussian enhancement.

\subsection{Underwater Spatial Gaussian}

\begin{figure}[t]
  \centering
  \includegraphics[width=1.0\linewidth]{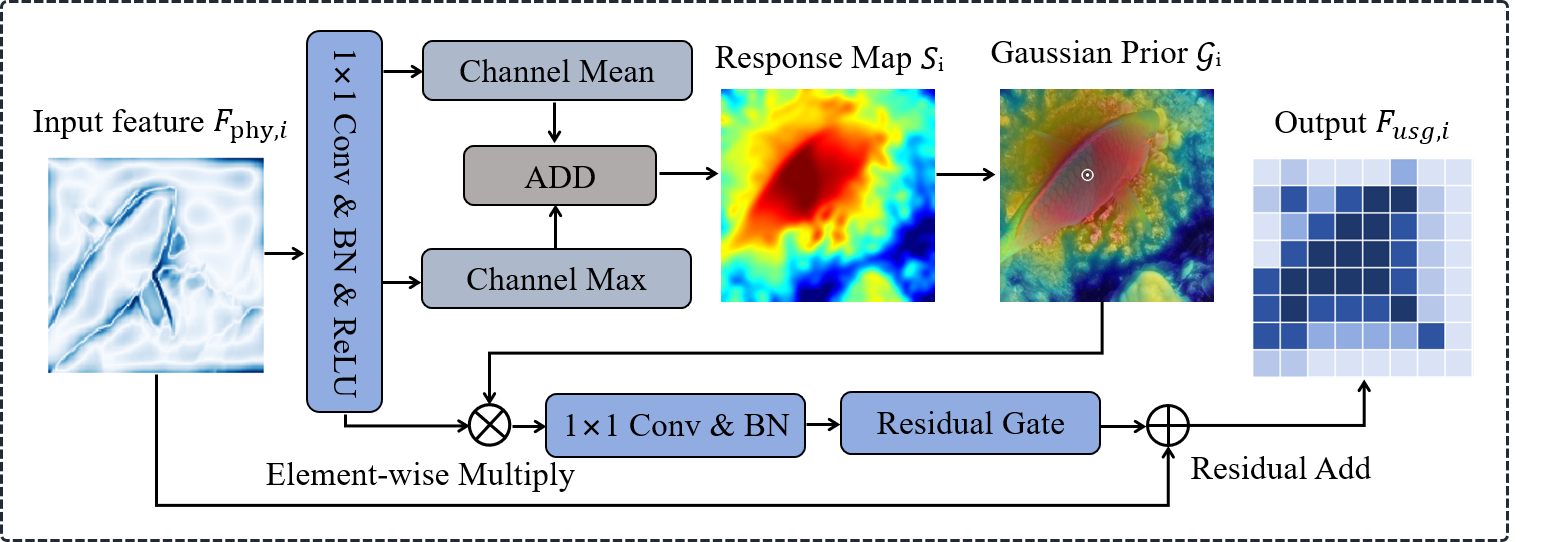}
  \caption{Overview of the Underwater Spatial Gaussian (USG) module. Given the physics-guided feature, USG computes a saliency response map by aggregating channel-wise mean and maximum responses, selects the strongest response as the saliency anchor, and constructs an object-centered Gaussian prior. The prior is then injected through a residual gate to enhance spatially coherent salient regions and suppress fragmented background responses.}
  \label{fig:fig4}
\end{figure}

As illustrated in Fig.~\ref{fig:fig4}, after UPP alleviates underwater-specific degradation, salient responses may still be fragmented due to cluttered backgrounds, weak boundaries, low contrast, and the irregular morphology of underwater organisms. Different from common land objects that often have relatively compact and regular shapes, many underwater salient targets, such as corals, crustaceans, and sea anemones, exhibit highly irregular contours, branch-like structures, or soft deformable boundaries. These characteristics easily lead to broken edges, incomplete salient regions, and discontinuous prediction masks. To address this issue, we propose an Underwater Spatial Gaussian module (USG), which aims to enhance object-centered saliency responses while preserving the spatial continuity of irregular aquatic targets.

Specifically, the physics-guided branch first estimates the pseudo-depth feature $F_{d,i}$ from the RGB feature $F_i$ according to Eq.~\eqref{eq:pseudo_depth}. Then, RGB and pseudo-depth features are interacted through a lightweight spatial-depth mixer and fed into UPP to obtain a physics-guided representation. Based on this representation, Spatial Gaussian Saliency Attention (SpatialGSA) selects the strongest response as the central saliency anchor and constructs a soft Gaussian prior around it. By starting from this reliable center anchor, USG softly propagates saliency confidence to neighboring object regions, which is helpful for recovering complete responses of coral-like irregular targets and alleviating edge fragmentation caused by underwater degradation.

We project RGB and pseudo-depth features into query and key embeddings, and generate a spatial-depth attention map by a convolutional mixer:
\begin{equation}
\begin{gathered}
Q_i = \phi_q(F_i), \quad K_i = \phi_k(F_{d,i}),\\[3pt]
A_i =
\sigma\left(
\mathrm{LN}\left(
\psi_{5\times5}^{dw}
\left(
\psi_{1\times1}([Q_i,K_i])
\right)
\right)
\right),
\end{gathered}
\label{eq:spatial_depth_attention}
\end{equation}
where $[Q_i,K_i]$ denotes channel-wise concatenation, $\psi_{1\times1}(\cdot)$ is used for channel mixing, $\psi_{5\times5}^{dw}(\cdot)$ denotes a depth-wise $5\times5$ convolution for local spatial interaction, and $\mathrm{LN}(\cdot)$ is layer normalization.

The spatial-depth attention map modulates RGB and pseudo-depth value features, which are then passed into the UPP branch to obtain a physics-guided representation:
\begin{equation}
\begin{gathered}
\hat{F}_{r,i}=F_i+\mathrm{DropPath}\left(\phi_r(A_i\odot \phi_{v_r}(F_i))\right),\\[3pt]
\hat{F}_{d,i}=F_{d,i}+\mathrm{DropPath}\left(\phi_d(A_i\odot \phi_{v_d}(F_{d,i}))\right),\\[3pt]
G_i = \mathcal{P}_{\mathrm{phy}}(\hat{F}_{r,i}, \hat{F}_{d,i}),
\end{gathered}
\label{eq:spatial_depth_modulation}
\end{equation}
where $\phi_{v_r}(\cdot)$ and $\phi_{v_d}(\cdot)$ denote value projections, $\phi_r(\cdot)$ and $\phi_d(\cdot)$ denote convolutional transformations, and $\mathcal{P}_{\mathrm{phy}}(\cdot)$ denotes the UPP operation described in Section~\ref{sec:upp}. Here, $G_i$ denotes the physics-guided feature produced by backscatter removal and attenuation compensation.

Based on the physics-guided representation $G_i$, SpatialGSA constructs an object-centered spatial prior. Specifically, a saliency response map is computed by combining channel-wise average and maximum responses, and the strongest response is selected as the saliency anchor:
\begin{equation}
\begin{gathered}
M_i = \mathrm{Avg}_c(\phi_p(G_i)) + \mathrm{Max}_c(\phi_p(G_i)),\\[3pt]
(c_y,c_x)=\arg\max_{(y,x)}M_i(y,x).
\end{gathered}
\label{eq:saliency_anchor}
\end{equation}
Then, a soft Gaussian prior is generated and injected into the guided feature through a learnable residual gate:
\begin{equation}
\begin{gathered}
\mathcal{G}_i(y,x)=
\exp\left(
-\frac{1}{2}
\left[
\frac{(y-c_y)^2}{\sigma_y^2}
+
\frac{(x-c_x)^2}{\sigma_x^2}
\right]
\right),\\[3pt]
S_i = G_i + \tanh(\eta)\cdot \phi_o(\phi_p(G_i)\odot \mathcal{G}_i),
\end{gathered}
\label{eq:gaussian_prior}
\end{equation}
where $\sigma_y$ and $\sigma_x$ are determined according to the feature-map size to provide a stable spatial spread of the Gaussian prior, $\eta$ is a learnable gate, and $\phi_o(\cdot)$ denotes an output projection. This residual design preserves the original physics-guided representation while softly expanding the most reliable salient response to neighboring object regions.

Finally, USG is integrated with the DMG branch through residual weighted fusion. Let $\mathcal{D}_{\mathrm{MG}}(F_i)$ denote the output of the dynamic multi-granularity branch. The final enhanced feature is computed by:
\begin{equation}
\begin{gathered}
\Delta_i^{m}= \mathcal{D}_{\mathrm{MG}}(F_i)-F_i,\quad
\Delta_i^{g}= \phi_g(S_i-F_i),\\[3pt]
F_i^{out} =
\alpha_i F_i +
\beta_i
\left(
\Delta_i^m+\tanh(\gamma_i)\Delta_i^g
\right),
\end{gathered}
\label{eq:residual_weighted_fusion}
\end{equation}
where $\phi_g(\cdot)$ is a projection layer for feature alignment, and $\alpha_i$, $\beta_i$, and $\gamma_i$ are learnable scale-aware fusion parameters. This design preserves the original RGB representation as the main path while adaptively injecting dynamic multi-granularity enhancement and underwater spatial Gaussian saliency guidance.

\subsection{Timestep-Adaptive Diffusion Transformer Bottleneck}

In the conditional diffusion framework, the denoising network needs to recover the salient mask from noisy samples at different timesteps. However, different denoising stages have different requirements. At early timesteps, the mask feature contains strong noise, and the network needs more global semantic reasoning to locate the salient object. At later timesteps, the noise level becomes lower, and the network should focus more on local boundary refinement. To improve the timestep adaptivity of the decoder, we introduce a lightweight DiT-style bottleneck after the fusion of conditional features and noisy mask features.

Let $F_c$ denote the fused conditional feature from the decoder and $F_t$ denote the noisy mask feature at timestep $t$. We first concatenate them and project the result into a unified feature space:
\begin{equation}
Z = \phi_{\mathrm{up}}\left([F_c, F_t]\right),
\end{equation}
where $[\cdot,\cdot]$ denotes channel-wise concatenation and $\phi_{\mathrm{up}}(\cdot)$ is a $1\times1$ projection. The timestep embedding is then used to generate adaptive modulation parameters:
\begin{equation}
(\Delta_a,s_a,g_a,\Delta_m,s_m,g_m)=\phi_t(t),
\end{equation}
where $\Delta_a$ and $s_a$ denote the shift and scale parameters for the self-attention branch, $g_a$ is the attention gate, while $\Delta_m$, $s_m$, and $g_m$ are the corresponding parameters for the MLP branch.

The timestep-adaptive refinement is formulated as:
\begin{equation}
\begin{gathered}
\hat{Z} = Z + g_a \cdot 
\mathrm{MSA}\left(\mathrm{LN}(Z)\odot(1+s_a)+\Delta_a\right),\\[4pt]
Z^{out} = \hat{Z} + g_m \cdot 
\mathrm{MLP}\left(\mathrm{LN}(\hat{Z})\odot(1+s_m)+\Delta_m\right),
\end{gathered}
\end{equation}
where $\mathrm{MSA}(\cdot)$ denotes multi-head self-attention, $\mathrm{MLP}(\cdot)$ denotes the feed-forward network, and $\mathrm{LN}(\cdot)$ denotes layer normalization. Following the AdaLN-Zero design, the last linear layer of $\phi_t(\cdot)$ is initialized to zero, so that the bottleneck starts as an identity-like residual block and gradually learns timestep-dependent refinement.

Through timestep-conditioned modulation, the DiT bottleneck enables the decoder to dynamically adjust its refinement behavior at different denoising stages. This is particularly important for underwater salient object detection, where salient objects often exhibit weak contrast and blurred boundaries. The self-attention branch captures long-range object dependencies, while the timestep-adaptive MLP branch refines local feature responses, jointly improving the completeness and boundary accuracy of the predicted salient masks.

\begin{table*}[t]
    \centering
    \caption{Quantitative comparison on the USOD10K and USOD benchmarks.}
    \vspace{3pt}
    \setlength{\tabcolsep}{5pt}
    \small
    \renewcommand{\arraystretch}{1.06}
    \resizebox{0.98\textwidth}{!}{%
    \begin{tabular}{c|c|c|cccc|cccc}
        \midrule
        \multirow{2}{*}{Method} & \multirow{2}{*}{Pub.} & \multirow{2}{*}{Backbone} & \multicolumn{4}{c|}{USOD10K} & \multicolumn{4}{c}{USOD} \\
        \cline{4-11}
        & & & \rule[-0.35ex]{0pt}{2.8ex}MAE $\downarrow$ & $E_{\phi}^{m}$ $\uparrow$ & $F_{\max}^{\beta}$ $\uparrow$ & $S_{\alpha}$ $\uparrow$
        & MAE $\downarrow$ & $E_{\phi}^{m}$ $\uparrow$ & $F_{\max}^{\beta}$ $\uparrow$ & $S_{\alpha}$ $\uparrow$ \\[1.2pt]
        \midrule
        HIDANet~\cite{wu2023hidanet} & TIP'23 & Res2Net-50 & 0.0247 & 0.9547 & 0.8976 & 0.9143 & 0.0459 & 0.9265 & 0.8954 & 0.8958 \\
        PICRNet~\cite{cong2023point} & ACM MM'23 & VGG-16 & 0.0233 & 0.9563 & 0.9045 & 0.9191 & 0.0435 & 0.9318 & 0.8945 & 0.9032 \\
        CATNet~\cite{sun2023catnet} & TMM'23 & Swin-B & 0.0232 & 0.8825 & 0.8669 & 0.9127 & 0.0433 & 0.9266 & 0.8959 & 0.8876 \\
        GeleNet~\cite{li2023salient} & TIP'23 & PVTv2 & 0.0229 & 0.9585 & 0.9139 & 0.9230 & 0.0428 & 0.9319 & 0.9061 & 0.9025 \\
        ISNet~\cite{zhu2024separate} & PR'24 & ResNet-50 & 0.0260 & 0.9504 & 0.8828 & 0.9157 & 0.0494 & 0.9214 & 0.8984 & 0.8944 \\
        VSCode~\cite{luo2024vscode} & CVPR'24 & Swin-S & 0.0221 & 0.9611 & 0.9281 & 0.9249 & 0.0438 & 0.9277 & 0.9146 & 0.8999 \\
        SPDE~\cite{jin2024underwater} & TCSVT'24 & Swin-T & 0.0203 & 0.9631 & 0.8996 & 0.9249 & 0.0444 & 0.9262 & 0.9010 & 0.9000 \\
        IGAN~\cite{mao2024generative} & TCSVT'24 & Swin-B & 0.0200 & 0.9654 & 0.9031 & 0.9206 & 0.0423 & 0.9341 & 0.9119 & 0.8990 \\
        DualSAM~\cite{zhang2024fantastic} & CVPR'24 & ViT-B & 0.0199 & 0.9638 & 0.9203 & 0.9243 & 0.0428 & 0.9324 & 0.8926 & 0.9014 \\
        SACNet~\cite{wang2024alignment} & TMM'24 & Swin-B & 0.0186 & 0.9672 & 0.9212 & 0.9280 & 0.0412 & 0.9337 & 0.9148 & 0.8995 \\
        SENet~\cite{hao2025simple} & TIP'25 & ViT-B & 0.0203 & 0.9612 & 0.9213 & 0.9253 & 0.0406 & 0.9326 & 0.9029 & 0.9028 \\
        DCNet~\cite{zhu2025dc} & PR'25 & Swin-B & 0.0202 & 0.9636 & 0.9152 & 0.9217 & 0.0449 & 0.9248 & 0.9047 & 0.8997 \\
        TC-USOD~\cite{hong2023usod10k} & TIP'25 & T2T-ViT & 0.0202 & 0.9636 & 0.9086 & 0.9217 & 0.0426 & 0.9331 & 0.8925 & 0.8994 \\
        FSCDiff~\cite{li2025fscdiff} & ACM MM'25 & PVTv2 & 0.0191 & 0.9641 & 0.9235 & 0.9286 & 0.0411 & 0.9289 & 0.9216 & 0.8994 \\
        Samba~\cite{he2025samba} & CVPR'25 & VMamba-S & 0.0187 & 0.9672 & 0.9339 & \underline{0.9312} & 0.0416 & 0.9302 & 0.9182 & \underline{0.9045} \\
        PLFRNet~\cite{han2025perceptual} & ESWA'25 & PVTv2 & \underline{0.0181} & \underline{0.9674} & 0.9319 & 0.9285 & 0.0422 & 0.9327 & 0.9151 & 0.9022 \\
        WaterFlow~\cite{li2026waterflow} & ICASSP'26 & PVTv2 & 0.0256 & 0.9529 & 0.8901 & 0.8987 & 0.0464 & 0.9268 & 0.8950 & 0.8793 \\
        CSSAM~\cite{cong2026breaking} & TPAMI'26 & Hiera-T & 0.0206 & 0.9614 & \underline{0.9356} & 0.9278 & 0.0434 & 0.9258 & \underline{0.9219} & 0.8987 \\
        SearchNet~\cite{zhou2026turbidity} & TIP'26 & PVTv2 & 0.0188 & 0.9672 & 0.9289 & 0.9263 & \textbf{0.0397} & \underline{0.9347} & 0.9197 & 0.9022 \\
        \midrule
        \textbf{DCGNet (Ours)} & -- & PVTv2 & \textbf{0.0173} & \textbf{0.9684} & \textbf{0.9370} & \textbf{0.9314} & \underline{0.0398} & \textbf{0.9363} & \textbf{0.9233} & \textbf{0.9048} \\
        \midrule
    \end{tabular}%
    }
    \label{tab:performance_comparison}
\end{table*}

\section{Experiments}

\subsection{Experimental Setup}

\textbf{Datasets.}
We conduct experiments on five benchmarks, including USOD~\cite{islam2020svam}, USOD10K~\cite{hong2023usod10k}, CSOD10K~\cite{cong2026breaking}, MAS3K~\cite{li2021mas3k}, and RMAS~\cite{fu2023masnet}. The USOD benchmark contains 300 underwater images with pixel-level annotations. The USOD10K benchmark contains 10,255 underwater images with pixel-level ground truths and depth maps, including 7,178 training images, 2,051 validation images, and 1,026 testing images. For USOD10K and USOD, we train DCGNet on the training set of USOD10K and evaluate it on USOD10K and USOD to verify both in-domain performance and cross-benchmark generalization. For CSOD10K, MAS3K, and RMAS, we follow their corresponding benchmark protocols and train and evaluate DCGNet on each benchmark separately. CSOD10K is a condition-constrained SOD benchmark that includes low-light, fog, rain, snow, underwater, reflection, blur, and overexposure scenes. Among them, low-light, fog, rain, snow, and underwater scenes share certain degradation similarities with underwater imaging, such as degraded visibility, low contrast, medium interference, and complex backgrounds. MAS3K and RMAS are adopted as other underwater-scene benchmarks to further evaluate the generalization ability of DCGNet under diverse underwater imaging conditions.

\textbf{Implementation Details.}
The proposed DCGNet is implemented using the publicly available PyTorch framework and two NVIDIA GeForce RTX 4090 GPUs. The backbone network is based on PVTv2~\cite{wang2022pvt}, with parameters pre-trained on ImageNet~\cite{deng2009imagenet} used for initialization. We train our model for 150 epochs with a batch size of 32. We use the Adam optimizer with an initial learning rate of $1\times10^{-5}$ and gradually reduce the learning rate to $1\times10^{-6}$ during training using CosineAnnealingLR. During both training and inference, the input image size is resized to 352$\times$352.

\textbf{Evaluation Criteria.}
We employ widely used salient object detection metrics for quantitative evaluation, including Mean Absolute Error (MAE)~\cite{perazzi2012saliency}, maximum F-measure ($F_{\max}^{\beta}$)~\cite{achanta2009frequency}, weighted F-measure ($F_{\omega}^{\beta}$)~\cite{margolin2014evaluate}, E-measure ($E_{\phi}^{m}$)~\cite{fan2018enhanced}, and S-measure ($S_{\alpha}$)~\cite{fan2017structure}. A lower MAE indicates better performance, while higher values of the other metrics indicate better performance. In all quantitative tables, the best results are highlighted in \textbf{bold}, and the second-best results are \underline{underlined}.
\subsection{Baselines}

\textbf{1) Methods on USOD10K and USOD.}
For the standard underwater salient object detection task, we compare DCGNet with nineteen representative methods on USOD10K and USOD, including GeleNet~\cite{li2023salient}, HIDANet~\cite{wu2023hidanet}, CATNet~\cite{sun2023catnet}, PICRNet~\cite{cong2023point}, ISNet~\cite{zhu2024separate}, DualSAM~\cite{zhang2024fantastic}, SACNet~\cite{wang2024alignment}, TC-USOD~\cite{hong2023usod10k}, DCNet~\cite{zhu2025dc}, SENet~\cite{hao2025simple}, SPDE~\cite{jin2024underwater}, IGAN~\cite{mao2024generative}, VSCode~\cite{luo2024vscode}, FSCDiff~\cite{li2025fscdiff}, Samba~\cite{he2025samba}, PLFRNet~\cite{han2025perceptual}, WaterFlow~\cite{li2026waterflow}, CSSAM~\cite{cong2026breaking}, and SearchNet~\cite{zhou2026turbidity}. These methods cover CNN-based, Transformer-based, SAM-based, RGB-D/underwater-specific, and generative saliency models, providing a comprehensive comparison under the conventional RGB-to-mask USOD evaluation protocol.

\textbf{2) Methods on CSOD10K.}
To evaluate robustness under constrained scenes that share degradation similarities with underwater imaging, we further report results on the CSOD10K benchmark~\cite{cong2026breaking}. We compare DCGNet with sixteen representative SOD/CSOD methods, including GateNet~\cite{zhao2020suppress}, MINet~\cite{pang2020multi}, LDF~\cite{wei2020label}, VST~\cite{liu2021visual}, EDN~\cite{wu2022edn}, TRACER~\cite{lee2022tracer}, ICON~\cite{zhuge2022salient}, SelfReformer~\cite{yun2022selfreformer}, BBRF~\cite{ma2023boosting}, SI-EDN~\cite{li2024size}, MDSAM~\cite{gao2024multi}, VSCode~\cite{luo2024vscode}, DC-Net~\cite{zhu2025dc}, FSCDiff~\cite{li2025fscdiff}, PLFRNet~\cite{han2025perceptual}, and CSSAM~\cite{cong2026breaking}.

\textbf{3) Methods on MAS3K and RMAS.}
For other underwater-scene benchmarks, we conduct additional comparisons on MAS3K~\cite{li2021mas3k} and RMAS~\cite{fu2023masnet}. Fourteen methods are selected, including OCENet~\cite{liu2022modeling}, ZoomNet~\cite{pang2022zoom}, MASNet~\cite{fu2023masnet}, SETR~\cite{zheng2021rethinking}, H2Former~\cite{he2023h2former}, SAM~\cite{kirillov2023segment}, SAM-Adapter~\cite{chen2023samadapter}, CPNet~\cite{hu2024cross}, DualSAM~\cite{zhang2024fantastic}, MAS-SAM~\cite{yan2024massam}, VSCode~\cite{luo2024vscode}, DCNet~\cite{zhu2025dc}, FSCDiff~\cite{li2025fscdiff}, and WaterFlow~\cite{li2026waterflow}. This comparison verifies whether the proposed model can generalize beyond USOD10K/USOD to more diverse underwater scenarios.

For fairness, all compared methods are evaluated under the corresponding RGB-to-mask saliency detection protocols. We use saliency maps provided by the authors when available, or generate predictions using released models or retrained models following the public codes and default settings. Since WaterDiffusion~\cite{chang2025waterdiffusion} follows a joint restoration-saliency formulation and uses additional restoration/medium-related supervision beyond the standard RGB-to-mask protocol, we discuss it as related generative work rather than include it in the direct quantitative comparison.

\subsection{Performance Comparison}

\begin{figure*}[t]
    \centering
    \includegraphics[width=1.0\linewidth]{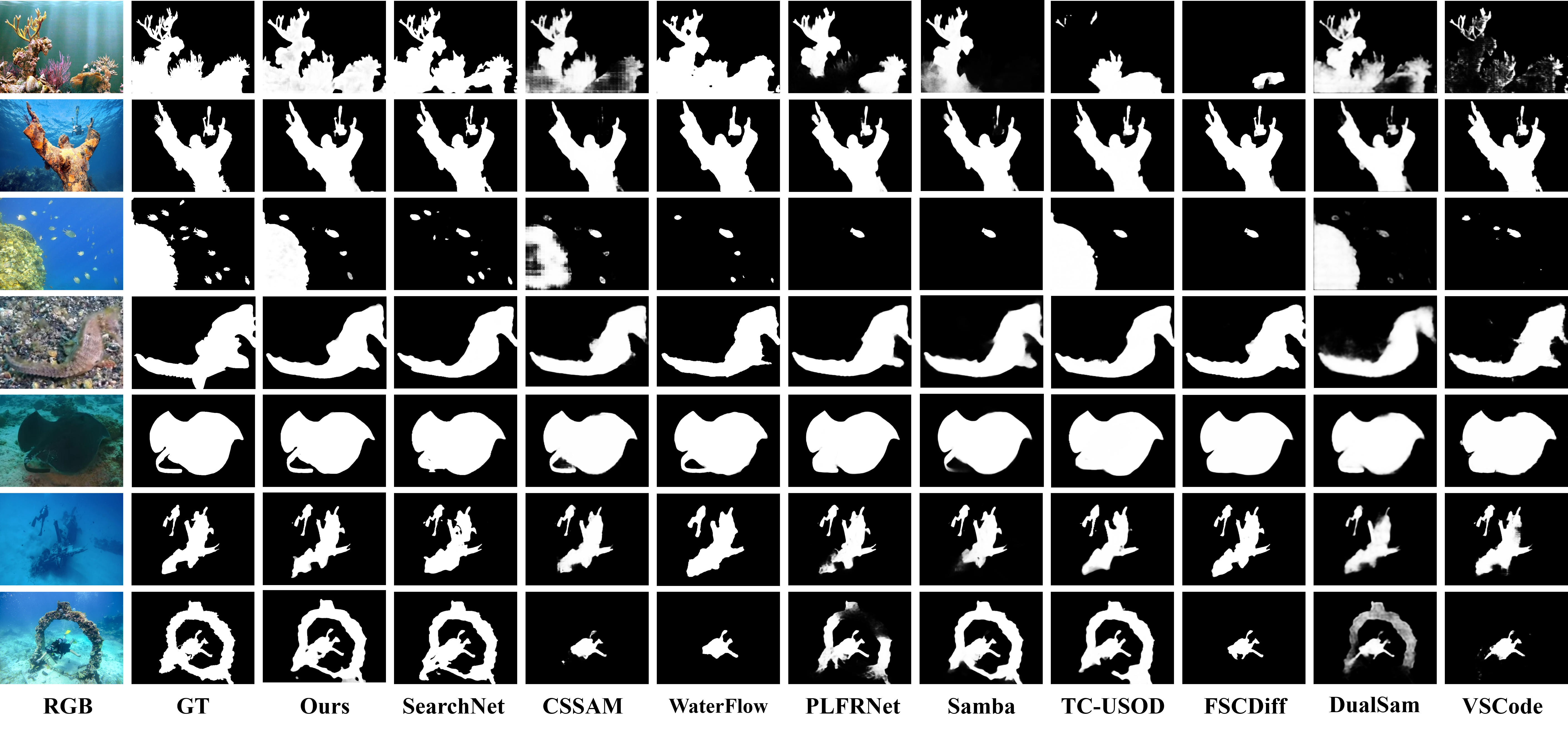}
    \caption{Qualitative comparison on the USOD10K~\cite{hong2023usod10k} and USOD~\cite{islam2020svam} benchmarks. The examples cover low contrast, severe color degradation, weak foreground contrast, and complex underwater backgrounds.}
    \label{fig:fig5}
\end{figure*}

\textbf{Quantitative and visual comparison on USOD10K and USOD.}
As shown in Table~\ref{tab:performance_comparison}, DCGNet achieves the best performance on all reported metrics of USOD10K and on most metrics of USOD. Specifically, on USOD10K, DCGNet obtains the lowest MAE of 0.0173 and the highest $E_{\phi}^{m}$, $F_{\max}^{\beta}$, and $S_{\alpha}$, indicating that the proposed degradation-aware condition construction and diffusion-based mask generation can produce accurate and structurally complete saliency maps. On the USOD benchmark, DCGNet achieves the best $E_{\phi}^{m}$, $F_{\max}^{\beta}$, and $S_{\alpha}$, while obtaining the second-best MAE, demonstrating strong cross-benchmark generalization to unseen underwater scenes. The visual comparison in Fig.~\ref{fig:fig5} further shows that DCGNet generates more complete salient regions and sharper object boundaries under low contrast, severe color degradation, and complex underwater backgrounds.

\begin{table}[t]
    \centering
    \caption{Quantitative comparison on the challenging CSOD10K benchmark.}
    \vspace{3pt}
    \setlength{\tabcolsep}{2.4pt}
    \scriptsize
    \resizebox{\columnwidth}{!}{%
    \begin{tabular}{c|c|c|cccc}
        \midrule
        Method & Pub. & Backbone & MAE $\downarrow$ & $F_{\max}^{\beta}$ $\uparrow$ & $S_{\alpha}$ $\uparrow$ & $E_{\phi}^{m}$ $\uparrow$ \\
        \midrule
        GateNet~\cite{zhao2020suppress} & ECCV'20 & ResNet-50 & 0.056 & 0.847 & 0.831 & 0.853 \\
        MINet~\cite{pang2020multi} & CVPR'20 & ResNet-50 & 0.051 & 0.831 & 0.824 & 0.856 \\
        LDF~\cite{wei2020label} & CVPR'20 & ResNet-50 & 0.049 & 0.853 & 0.837 & 0.872 \\
        VST~\cite{liu2021visual} & ICCV'21 & T2T-ViT & 0.047 & 0.835 & 0.843 & 0.871 \\
        EDN~\cite{wu2022edn} & TIP'22 & ResNet-50 & 0.053 & 0.819 & 0.829 & 0.863 \\
        TRACER~\cite{lee2022tracer} & AAAI'22 & EfficientNet-B5 & 0.041 & 0.849 & 0.855 & 0.899 \\
        ICON~\cite{zhuge2022salient} & TPAMI'22 & Swin-B & 0.041 & 0.865 & 0.862 & \underline{0.904} \\
        SelfReformer~\cite{yun2022selfreformer} & TMM'23 & PVTv2 & 0.047 & 0.857 & 0.856 & 0.886 \\
        BBRF~\cite{ma2023boosting} & TIP'23 & Swin-B & 0.043 & 0.824 & 0.846 & 0.894 \\
        SI-EDN~\cite{li2024size} & ICML'24 & ResNet-50 & 0.062 & 0.818 & 0.811 & 0.841 \\
        MDSAM~\cite{gao2024multi} & ACM MM'24 & ViT-B & 0.042 & \underline{0.874} & 0.868 & 0.895 \\
        VSCode~\cite{luo2024vscode} & CVPR'24 & Swin-S & 0.040 & 0.833 & 0.862 & 0.896 \\
        FSCDiff~\cite{li2025fscdiff} & ACM MM'25 & PVTv2 & 0.063 & 0.850 & 0.829 & 0.873 \\
        DC-Net~\cite{zhu2025dc} & PR'25 & Swin-B & 0.057 & 0.824 & 0.821 & 0.857 \\
        PLFRNet~\cite{han2025perceptual} & ESWA'25 & PVTv2 & 0.040 & 0.868 & 0.863 & 0.899 \\
        CSSAM~\cite{cong2026breaking} & TPAMI'26 & Hiera-T & \underline{0.040} & 0.870 & \underline{0.871} & 0.903 \\
        \midrule
        \textbf{DCGNet} & -- & PVTv2 & \textbf{0.038} & \textbf{0.876} & \textbf{0.878} & \textbf{0.910} \\
        \midrule
    \end{tabular}%
    }
    \label{tab:csod10k_comparison}
\end{table}

\begin{figure}[t]
    \centering
    \includegraphics[width=1.0\linewidth]{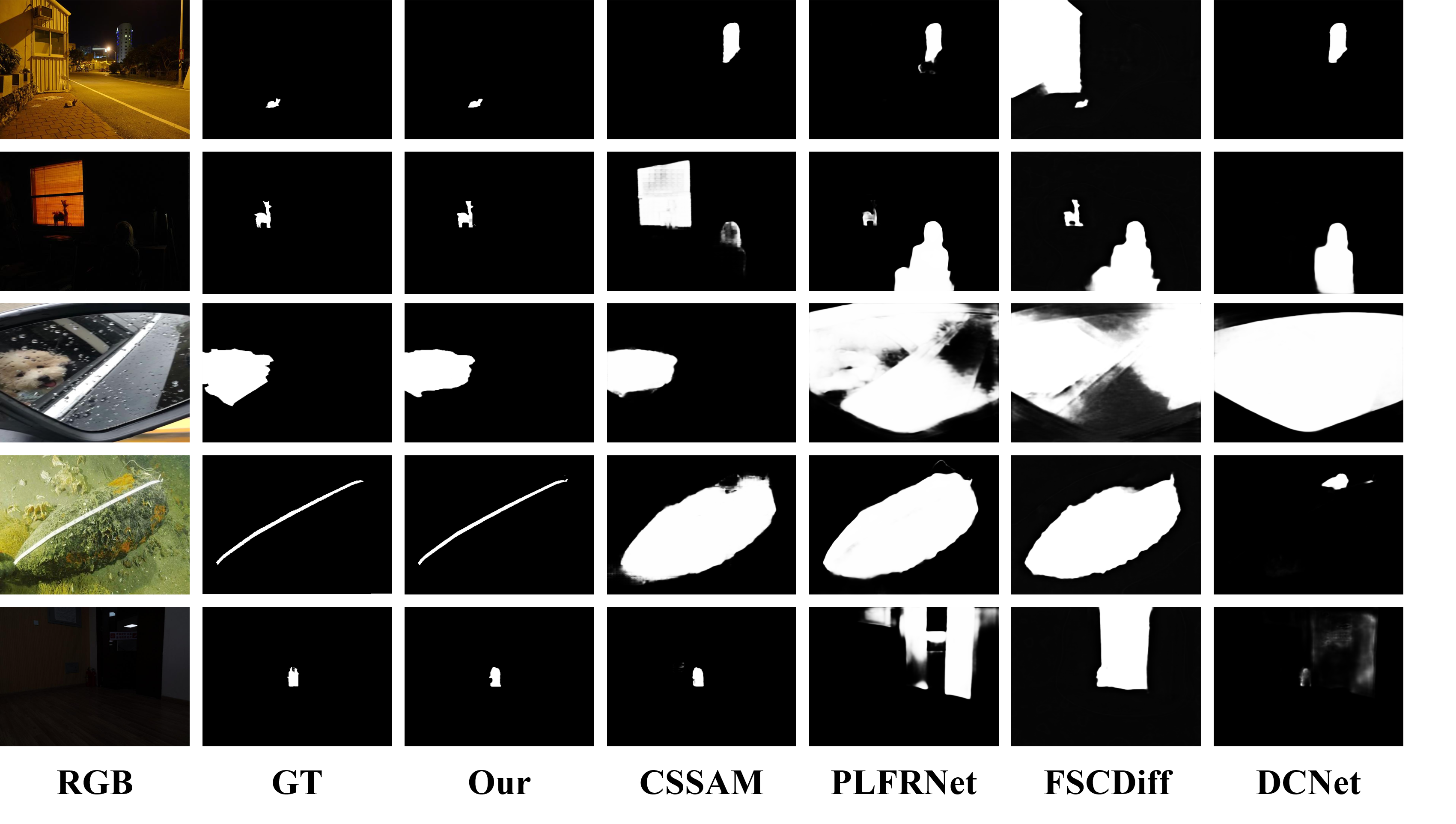}
    \caption{Qualitative comparison on the CSOD10K benchmark. The examples illustrate representative constrained scenes with degraded visibility, low-light interference, and cluttered backgrounds.}
    \label{fig:fig6}
\end{figure}

\textbf{Quantitative and visual comparison on CSOD10K.}
Table~\ref{tab:csod10k_comparison} reports the results on the challenging CSOD10K benchmark, where DCGNet achieves the best MAE, $F_{\max}^{\beta}$, $S_{\alpha}$, and $E_{\phi}^{m}$. Since CSOD10K contains rain, snow, fog, and low-light scenes with degradation characteristics similar to underwater imaging, these results show that DMG, UPP, and USG are effective under restricted and cluttered visual conditions. Visual comparison in Fig.~\ref{fig:fig6} further demonstrates that DCGNet produces more complete salient regions and suppresses interference.

\begin{table}[t]
    \centering
    \caption{Quantitative comparison on the MAS3K and RMAS benchmarks.}
    \vspace{3pt}
    \setlength{\tabcolsep}{1.2pt}
    \scriptsize
    \renewcommand{\arraystretch}{1.06}
    \resizebox{\columnwidth}{!}{%
    \begin{tabular}{c|c|cccc|cccc}
        \midrule
        \multirow{2}{*}{Method} & \multirow{2}{*}{Pub.} & \multicolumn{4}{c|}{MAS3K~\cite{li2021mas3k}} & \multicolumn{4}{c}{RMAS~\cite{fu2023masnet}} \\
        \cline{3-10}
        & & \rule[-0.35ex]{0pt}{2.8ex}$S_{\alpha}$ $\uparrow$ & $F_{\omega}^{\beta}$ $\uparrow$ & $E_{\phi}^{m}$ $\uparrow$ & MAE $\downarrow$
        & $S_{\alpha}$ $\uparrow$ & $F_{\omega}^{\beta}$ $\uparrow$ & $E_{\phi}^{m}$ $\uparrow$ & MAE $\downarrow$ \\[1.2pt]
        \midrule
        SETR~\cite{zheng2021rethinking} & CVPR'21 & 0.855 & 0.789 & 0.917 & 0.030 & 0.818 & 0.747 & 0.933 & 0.028 \\
        OCENet~\cite{liu2022modeling} & WACV'22 & 0.824 & 0.703 & 0.868 & 0.052 & 0.836 & 0.752 & 0.900 & 0.030 \\
        ZoomNet~\cite{pang2022zoom} & CVPR'22 & 0.862 & 0.780 & 0.898 & 0.032 & 0.855 & 0.795 & 0.915 & 0.022 \\
        SAM~\cite{kirillov2023segment} & ICCV'23 & 0.763 & 0.656 & 0.807 & 0.059 & 0.697 & 0.534 & 0.790 & 0.053 \\
        SAM-Adapter~\cite{chen2023samadapter} & ICCVW'23 & 0.847 & 0.782 & 0.914 & 0.033 & 0.816 & 0.752 & 0.927 & 0.027 \\
        MASNet~\cite{fu2023masnet} & IJCAI'23 & 0.864 & 0.788 & 0.906 & 0.032 & 0.862 & 0.801 & 0.920 & 0.024 \\
        H2Former~\cite{he2023h2former} & TMI'23 & 0.865 & 0.810 & 0.925 & 0.028 & 0.844 & 0.799 & 0.931 & 0.023 \\
        CPNet~\cite{hu2024cross} & IJCV'24 & 0.869 & 0.805 & 0.918 & 0.028 & 0.837 & 0.791 & 0.939 & 0.024 \\
        MAS-SAM~\cite{yan2024massam} & IJCAI'24 & \underline{0.887} & \underline{0.840} & 0.938 & 0.025 & 0.865 & 0.819 & \underline{0.948} & \underline{0.021} \\
        VSCode~\cite{luo2024vscode} & CVPR'24 & 0.856 & 0.839 & \underline{0.947} & \underline{0.023} & \underline{0.875} & \underline{0.822} & 0.933 & 0.027 \\
        DualSAM~\cite{zhang2024fantastic} & CVPR'24 & 0.884 & 0.838 & 0.933 & \underline{0.023} & 0.860 & 0.812 & 0.944 & 0.022 \\
        FSCDiff~\cite{li2025fscdiff} & ACM MM'25 & 0.722 & 0.698 & 0.805 & 0.064 & 0.856 & 0.819 & 0.947 & 0.023 \\
        DCNet~\cite{zhu2025dc} & PR'25 & 0.829 & 0.774 & 0.933 & 0.042 & 0.838 & 0.804 & 0.945 & 0.029 \\
        WaterFlow~\cite{li2026waterflow} & ICASSP'26 & 0.810 & 0.813 & 0.882 & 0.051 & 0.838 & 0.813 & 0.921 & 0.024 \\
        \midrule
        \textbf{DCGNet} & -- & \textbf{0.902} & \textbf{0.861} & \textbf{0.952} & \textbf{0.019} & \textbf{0.880} & \textbf{0.839} & \textbf{0.961} & \textbf{0.018} \\
        \midrule
    \end{tabular}%
    }
    \label{tab:other_underwater_comparison}
\end{table}

\begin{figure}[t]
    \centering
    \includegraphics[width=1.0\linewidth]{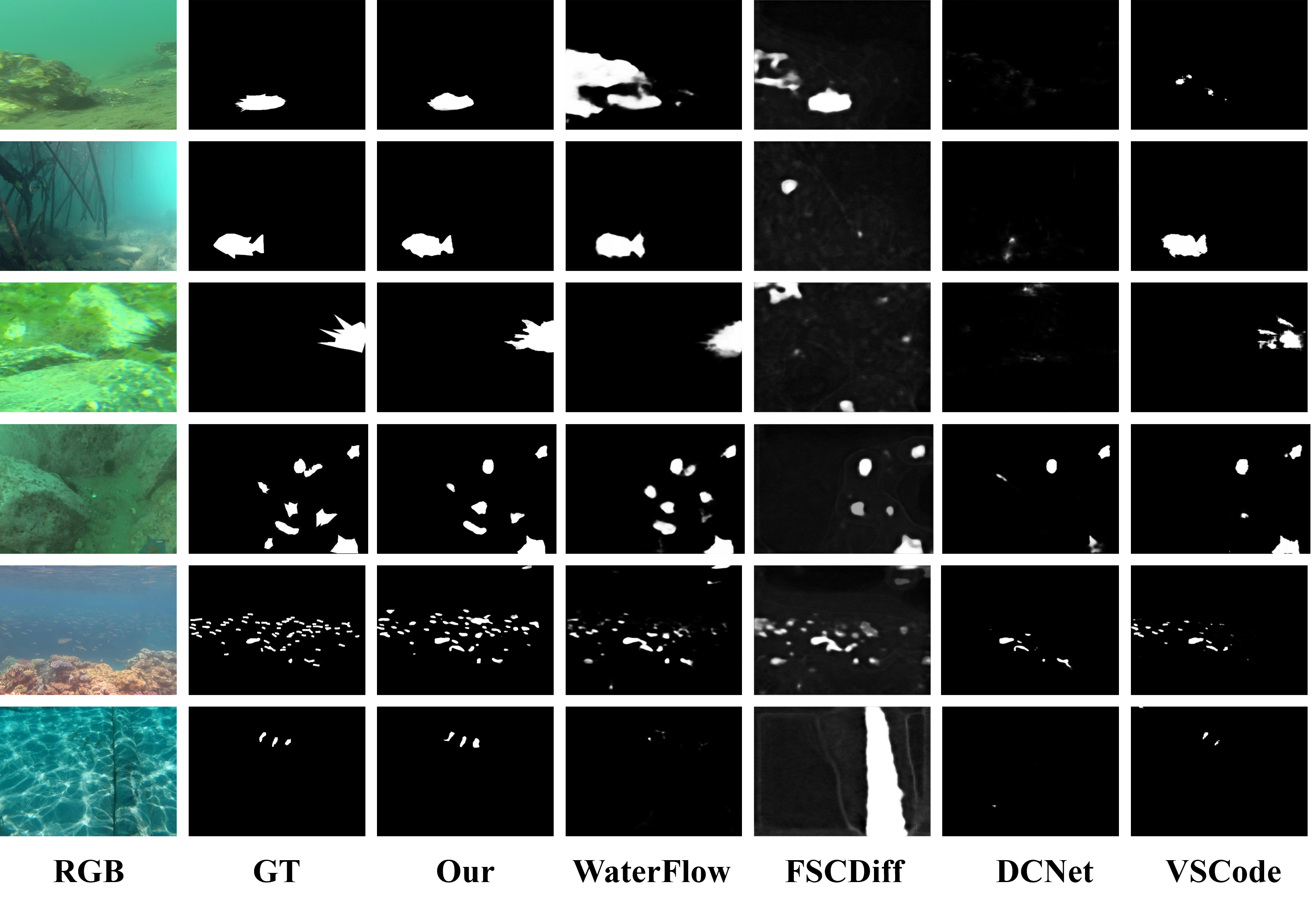}
    \caption{Qualitative comparison on the MAS3K and RMAS benchmarks. The examples include diverse marine animals and complex underwater backgrounds, evaluating cross-benchmark robustness.}
    \label{fig:fig7}
\end{figure}

\textbf{Quantitative and visual comparison on MAS3K and RMAS.}
Table~\ref{tab:other_underwater_comparison} presents additional comparisons on MAS3K and RMAS. DCGNet consistently achieves the best results on both benchmarks, including lower MAE and higher structure-measure, weighted F-measure, and enhanced-measure scores. The relatively unstable performance of existing generative underwater salient object detection models on these additional benchmarks also indicates that their cross-benchmark generalization ability still remains insufficient. Visual comparison in Fig.~\ref{fig:fig7} shows that DCGNet better preserves salient object structures across different underwater scenes, further confirming its generality across diverse underwater benchmarks and annotation protocols.

\subsection{Ablation Studies}

\begin{table}[t]
    \centering
    \caption{Ablation study of the proposed components on the USOD10K benchmark.}
    \vspace{3pt}
    \setlength{\tabcolsep}{2.0pt}
    \scriptsize
    \resizebox{\columnwidth}{!}{%
    \begin{tabular}{c|cccc|cccc}
        \midrule
        ID & DMG & UPP & USG & TAD & MAE $\downarrow$ & $E_{\phi}^{m}$ $\uparrow$ & $F_{\max}^{\beta}$ $\uparrow$ & $S_{\alpha}$ $\uparrow$ \\
        \midrule
        A & $\checkmark$ & $\checkmark$ & -- & -- & 0.0188 & 0.9674 & 0.9271 & 0.9253 \\
        B & $\checkmark$ & -- & $\checkmark$ & -- & 0.0193 & 0.9639 & 0.9263 & 0.9224 \\
        C & $\checkmark$ & -- & -- & $\checkmark$ & 0.0190 & 0.9645 & 0.9226 & 0.9251 \\
        D & -- & $\checkmark$ & $\checkmark$ & -- & 0.0208 & 0.9634 & 0.9181 & 0.9219 \\
        E & -- & $\checkmark$ & -- & $\checkmark$ & 0.0224 & 0.9598 & 0.9003 & 0.9095 \\
        F & -- & -- & $\checkmark$ & $\checkmark$ & 0.0235 & 0.9559 & 0.8997 & 0.9050 \\
        G & -- & $\checkmark$ & $\checkmark$ & $\checkmark$ & 0.0200 & 0.9650 & 0.9345 & 0.9281 \\
        H & $\checkmark$ & -- & $\checkmark$ & $\checkmark$ & 0.0187 & 0.9667 & \underline{0.9366} & 0.9292 \\
        I & $\checkmark$ & $\checkmark$ & -- & $\checkmark$ & 0.0181 & 0.9674 & 0.9358 & 0.9301 \\
        J & $\checkmark$ & $\checkmark$ & $\checkmark$ & -- & \underline{0.0177} & \underline{0.9680} & 0.9362 & \underline{0.9303} \\
        \midrule
        K & $\checkmark$ & $\checkmark$ & $\checkmark$ & $\checkmark$ & \textbf{0.0173} & \textbf{0.9684} & \textbf{0.9370} & \textbf{0.9314} \\
        \midrule
    \end{tabular}%
    }
    \label{tab:ablation_study}
\end{table}

\begin{figure}[t]
    \centering
    \includegraphics[width=1\linewidth]{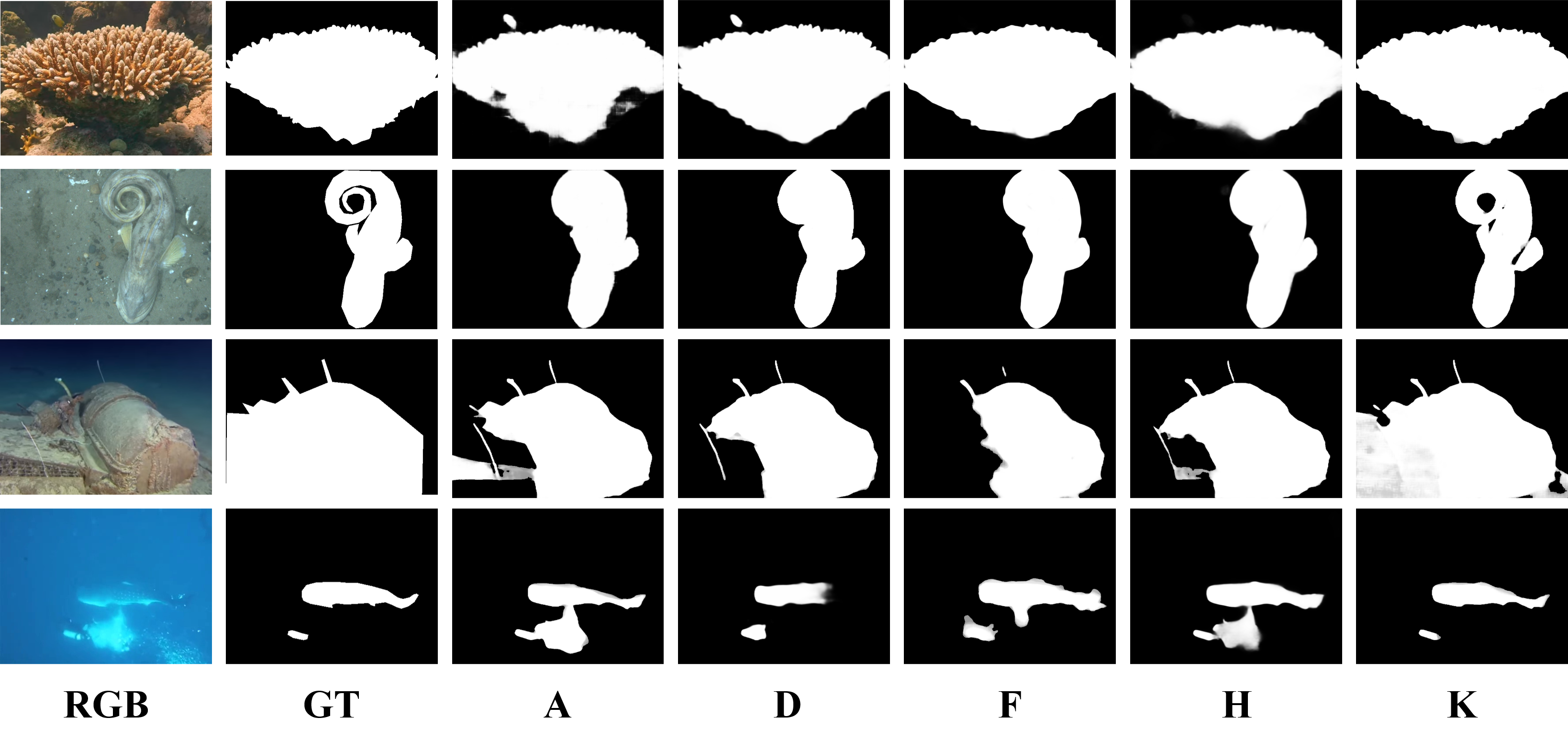}
    \caption{Qualitative ablation results on the USOD10K benchmark, showing the visual contribution of each proposed component.}
    \label{fig:fig8}
\end{figure}

In this section, we validate the contribution of each core component of DCGNet through ablation experiments on the USOD10K benchmark. Table~\ref{tab:ablation_study} reports the component-wise results. The ID column is used to match each configuration with the corresponding visual comparison in Fig.~\ref{fig:fig8}. Here, DMG denotes Dynamic Multi-Granularity, UPP denotes Underwater Physics-Prior, USG denotes Underwater Spatial Gaussian, and TAD denotes the timestep-adaptive DiT bottleneck.

\textbf{Effectiveness of DMG.}
Comparing the configurations with and without DMG shows that DMG consistently improves the overall performance. When DMG is introduced together with UPP, USG, and TAD, the full model achieves the best MAE, $E_{\phi}^{m}$, $F_{\max}^{\beta}$, and $S_{\alpha}$. This confirms that multi-granularity perception is important for capturing both coarse global context and fine local boundary details in underwater scenes.

\textbf{Effectiveness of UPP.}
UPP contributes to degradation-aware feature modeling by using pseudo-depth guidance to approximate underwater backscatter removal and attenuation compensation. The ablation results show that configurations containing UPP generally obtain stronger performance than their counterparts without UPP, especially when UPP is combined with DMG and USG. This demonstrates that physics-guided feature enhancement provides a cleaner condition for diffusion-based saliency generation, reducing the risk that underwater color distortion and backscatter noise are repeatedly propagated during iterative denoising.

\textbf{Effectiveness of USG.}
USG further enhances object-centered salient regions by constructing a soft spatial Gaussian prior from the strongest guided response. The improvements brought by USG indicate that spatial prior modeling is beneficial for suppressing cluttered background responses and recovering complete salient regions, especially for irregular underwater objects with weak boundaries. For diffusion generation, this object-centered prior helps the denoising trajectory maintain spatial consistency instead of drifting toward fragmented local responses.

\textbf{Effectiveness of TAD.}
The timestep-adaptive DiT bottleneck also improves the final performance by enabling the denoising decoder to adapt its refinement behavior at different diffusion timesteps. The full model with all four components achieves the best result, demonstrating that condition enhancement and timestep-aware generative refinement are complementary.

\section{Conclusion}

In this paper, we propose the Degradation-aware Conditional Generation Network (DCGNet) for underwater salient object detection. The DMG module enhances boundary extraction and captures multi-scale features, thereby improving the conditional input of the diffusion model. The UPP module leverages physics-guided priors to restore degraded RGB features by modeling light attenuation and backscatter effects. Based on the physics-guided representation, USG introduces SpatialGSA-based Gaussian saliency priors to enhance object-centered regions and suppress cluttered backgrounds. In addition, a lightweight timestep-adaptive DiT bottleneck improves global localization and boundary refinement at different diffusion timesteps. Experiments on multiple underwater benchmarks show that DCGNet achieves state-of-the-art performance and alleviates color distortion, boundary blurring, low contrast, and fragmented saliency responses.

\bibliographystyle{IEEEtran}
\bibliography{references}

@String(ICASSP=	{ICASSP})

@String(AAAI = {AAAI})

@inproceedings{zhu2023conditional,
  title={Conditional text image generation with diffusion models},
  author={Zhu, Yuanzhi and Li, Zhaohai and Wang, Tianwei and He, Mengchao and Yao, Cong},
  booktitle={Proceedings of the IEEE/CVF Conference on Computer Vision and Pattern Recognition},
  pages={14235--14245},
  year={2023}
}

@article{moser2024diffusion,
  title={Diffusion models, image super-resolution, and everything: A survey},
  author={Moser, Brian B and Shanbhag, Arundhati S and Raue, Federico and Frolov, Stanislav and Palacio, Sebastian and Dengel, Andreas},
  journal={IEEE Transactions on Neural Networks and Learning Systems},
  year={2024},
  publisher={IEEE}
}

@inproceedings{chen2024camodiffusion,
  title={CamoDiffusion: Camouflaged object detection via conditional diffusion models},
  author={Chen, Zhongxi and Sun, Ke and Lin, Xianming},
  booktitle={Proceedings of the AAAI Conference on Artificial Intelligence},
  volume={38},
  number={2},
  pages={1272--1280},
  year={2024}
}

@article{sun2025conditional,
  title={Conditional diffusion models for camouflaged and salient object detection},
  author={Sun, Ke and Chen, Zhongxi and Lin, Xianming and Sun, Xiaoshuai and Liu, Hong and Ji, Rongrong},
  journal={IEEE Transactions on Pattern Analysis and Machine Intelligence},
  year={2025},
  publisher={IEEE}
}

@inproceedings{wu2024medsegdiff,
  title={Medsegdiff-v2: Diffusion-based medical image segmentation with transformer},
  author={Wu, Junde and Ji, Wei and Fu, Huazhu and Xu, Min and Jin, Yueming and Xu, Yanwu},
  booktitle={Proceedings of the AAAI Conference on Artificial Intelligence},
  volume={38},
  number={6},
  pages={6030--6038},
  year={2024}
}

@inproceedings{mei2024codi,
  title={Codi: Conditional diffusion distillation for higher-fidelity and faster image generation},
  author={Mei, Kangfu and Delbracio, Mauricio and Talebi, Hossein and Tu, Zhengzhong and Patel, Vishal M and Milanfar, Peyman},
  booktitle={Proceedings of the IEEE/CVF Conference on Computer Vision and Pattern Recognition},
  pages={9048--9058},
  year={2024}
}

@inproceedings{wang2025lsnet,
  title={LSNet: See Large, Focus Small},
  author={Wang, Ao and Chen, Hui and Lin, Zijia and Han, Jungong and Ding, Guiguang},
  booktitle={Proceedings of the Computer Vision and Pattern Recognition Conference},
  pages={9718--9729},
  year={2025}
}

@inproceedings{shi2024transnext,
  title={Transnext: Robust foveal visual perception for vision transformers},
  author={Shi, Dai},
  booktitle={Proceedings of the IEEE/CVF Conference on Computer Vision and Pattern Recognition},
  pages={17773--17783},
  year={2024}
}

@inproceedings{yang2021focal,
  title={Focal Attention for Long-Range Interactions in Vision Transformers},
  author={Yang, Jianwei and Li, Chunyuan and Zhang, Pengchuan and Dai, Xiyang and Xiao, Bin and Yuan, Lu and Gao, Jianfeng},
  booktitle={Advances in Neural Information Processing Systems},
  volume={34},
  pages={30008--30022},
  year={2021}
}

@inproceedings{achanta2009frequency,
  title={Frequency-tuned salient region detection},
  author={Achanta, Radhakrishna and Hemami, Sheila and Estrada, Francisco and Susstrunk, Sabine},
  booktitle={2009 IEEE Conference on Computer Vision and Pattern Recognition},
  pages={1597--1604},
  year={2009},
  organization={IEEE}
}

@inproceedings{akkaynak2019sea,
  title={Sea-thru: A method for removing water from underwater images},
  author={Akkaynak, Derya and Treibitz, Tali},
  booktitle={Proceedings of the IEEE/CVF Conference on Computer Vision and Pattern Recognition},
  pages={1682--1691},
  year={2019}
}

@inproceedings{cong2023point,
  title={Point-aware interaction and CNN-induced refinement network for RGB-D salient object detection},
  author={Cong, Runmin and Liu, Hongyu and Zhang, Chen and Zhang, Wei and Zheng, Feng and Song, Ran and Kwong, Sam},
  booktitle={Proceedings of the 31st ACM International Conference on Multimedia},
  pages={406--416},
  year={2023}
}

@inproceedings{deng2009imagenet,
  title={Imagenet: A large-scale hierarchical image database},
  author={Deng, Jia and Dong, Wei and Socher, Richard and Li, Li-Jia and Li, Kai and Fei-Fei, Li},
  booktitle={2009 IEEE Conference on Computer Vision and Pattern Recognition},
  pages={248--255},
  year={2009},
  organization={IEEE}
}

@inproceedings{fan2017structure,
  title={Structure-measure: A new way to evaluate foreground maps},
  author={Fan, Deng-Ping and Cheng, Ming-Ming and Liu, Yun and Li, Tao and Borji, Ali},
  booktitle={Proceedings of the IEEE International Conference on Computer Vision},
  pages={4548--4557},
  year={2017}
}

@inproceedings{fan2018enhanced,
  title={Enhanced-Alignment Measure for Binary Foreground Map Evaluation},
  author={Fan, Deng-Ping and Gong, Cheng and Cao, Yang and Ren, Bo and Cheng, Ming-Ming and Borji, Ali},
  booktitle={Proceedings of the Twenty-Seventh International Joint Conference on Artificial Intelligence},
  pages={698--704},
  year={2018},
  doi={10.24963/ijcai.2018/97}
}

@article{hong2023usod10k,
  title={Usod10k: a new benchmark dataset for underwater salient object detection},
  author={Hong, Lin and Wang, Xin and Zhang, Gan and Zhao, Ming},
  journal={IEEE Transactions on Image Processing},
  volume={34},
  pages={1602--1615},
  year={2023},
  publisher={IEEE}
}

@article{hu2024cross,
  title={Cross-modal fusion and progressive decoding network for RGB-D salient object detection},
  author={Hu, Xihang and Sun, Fuming and Sun, Jing and Wang, Fasheng and Li, Haojie},
  journal={International Journal of Computer Vision},
  volume={132},
  number={8},
  pages={3067--3085},
  year={2024},
  publisher={Springer}
}

@inproceedings{islam2020svam,
  title={{SVAM}: Saliency-Guided Visual Attention Modeling by Autonomous Underwater Robots},
  author={Islam, Md Jahidul and Wang, Ruobing and Sattar, Junaed},
  booktitle={Robotics: Science and Systems},
  year={2022},
  address={New York, NY, USA},
  doi={10.15607/RSS.2022.XVIII.048}
}

@article{jin2024underwater,
  title={Underwater Salient Object Detection via Dual-Stage Self-Paced Learning and Depth Emphasis},
  author={Jin, Jianhui and Jiang, Qiuping and Wu, Qingyuan and Xu, Binwei and Cong, Runmin},
  journal={IEEE Transactions on Circuits and Systems for Video Technology},
  volume={35},
  number={3},
  pages={2147--2160},
  year={2025},
  doi={10.1109/TCSVT.2024.3491907},
  publisher={IEEE}
}

@inproceedings{liu2021visual,
  title={Visual saliency transformer},
  author={Liu, Nian and Zhang, Ni and Wan, Kaiyuan and Shao, Ling and Han, Junwei},
  booktitle={Proceedings of the IEEE/CVF International Conference on Computer Vision},
  pages={4722--4732},
  year={2021}
}

@article{sun2023catnet,
  title={CATNet: A cascaded and aggregated transformer network for RGB-D salient object detection},
  author={Sun, Fuming and Ren, Peng and Yin, Bowen and Wang, Fasheng and Li, Haojie},
  journal={IEEE Transactions on Multimedia},
  volume={26},
  pages={2249--2262},
  year={2023},
  publisher={IEEE}
}

@article{wang2022pvt,
  title={Pvt v2: Improved baselines with pyramid vision transformer},
  author={Wang, Wenhai and Xie, Enze and Li, Xiang and Fan, Deng-Ping and Song, Kaitao and Liang, Ding and Lu, Tong and Luo, Ping and Shao, Ling},
  journal={Computational Visual Media},
  volume={8},
  number={3},
  pages={415--424},
  year={2022},
  publisher={TUP}
}

@article{yuan2025if,
  title={IF-USOD: Multimodal information fusion interactive feature enhancement architecture for underwater salient object detection},
  author={Yuan, Genji and Song, Jintao and Li, Jinjiang},
  journal={Information Fusion},
  volume={117},
  pages={102806},
  year={2025},
  publisher={Elsevier}
}

@inproceedings{zhang2024fantastic,
  title={Fantastic animals and where to find them: Segment any marine animal with dual sam},
  author={Zhang, Pingping and Yan, Tianyu and Liu, Yang and Lu, Huchuan},
  booktitle={Proceedings of the IEEE/CVF Conference on Computer Vision and Pattern Recognition},
  pages={2578--2587},
  year={2024}
}

@article{zhang2023underwater,
  title={Underwater image enhancement via weighted wavelet visual perception fusion},
  author={Zhang, Weidong and Zhou, Ling and Zhuang, Peixian and Li, Guohou and Pan, Xipeng and Zhao, Wenyi and Li, Chongyi},
  journal={IEEE Transactions on Circuits and Systems for Video Technology},
  volume={34},
  number={4},
  pages={2469--2483},
  year={2023},
  publisher={IEEE}
}

@inproceedings{he2025samba,
  title={Samba: A Unified Mamba-based Framework for General Salient Object Detection},
  author={He, Jiahao and Fu, Keren and Liu, Xiaohong and Zhao, Qijun},
  booktitle={Proceedings of the Computer Vision and Pattern Recognition Conference},
  pages={25314--25324},
  year={2025}
}

@article{zhu2025dc,
  title={Dc-net: Divide-and-conquer for salient object detection},
  author={Zhu, Jiayi and Qin, Xuebin and Elsaddik, Abdulmotaleb},
  journal={Pattern Recognition},
  volume={157},
  pages={110903},
  year={2025},
  publisher={Elsevier}
}

@article{han2025perceptual,
  title={Perceptual localization and focus refinement network for RGB-D salient object detection},
  author={Han, Jinyu and Wang, Mengyin and Wu, Weiyi and Jia, Xu},
  journal={Expert Systems with Applications},
  volume={259},
  pages={125278},
  year={2025},
  publisher={Elsevier}
}

@inproceedings{perazzi2012saliency,
  title={Saliency filters: Contrast based filtering for salient region detection},
  author={Perazzi, Federico and Kr{\"a}henb{\"u}hl, Philipp and Pritch, Yael and Hornung, Alexander},
  booktitle={2012 IEEE Conference on Computer Vision and Pattern Recognition},
  pages={733--740},
  year={2012},
  organization={IEEE}
}

@inproceedings{margolin2014evaluate,
  title={How to evaluate foreground maps?},
  author={Margolin, Ran and Zelnik-Manor, Lihi and Tal, Ayellet},
  booktitle={Proceedings of the IEEE Conference on Computer Vision and Pattern Recognition},
  pages={248--255},
  year={2014}
}

@article{li2023salient,
  title={Salient object detection in optical remote sensing images driven by transformer},
  author={Li, Gongyang and Bai, Zhen and Liu, Zhi and Zhang, Xinpeng and Ling, Haibin},
  journal={IEEE Transactions on Image Processing},
  volume={32},
  pages={5257--5269},
  year={2023},
  publisher={IEEE}
}

@article{wu2023hidanet,
  title={Hidanet: Rgb-d salient object detection via hierarchical depth awareness},
  author={Wu, Zongwei and Allibert, Guillaume and Meriaudeau, Fabrice and Ma, Chao and Demonceaux, C{\'e}dric},
  journal={IEEE Transactions on Image Processing},
  volume={32},
  pages={2160--2173},
  year={2023},
  publisher={IEEE}
}

@article{hao2025simple,
  title={A simple yet effective network based on vision transformer for camouflaged object and salient object detection},
  author={Hao, Chao and Yu, Zitong and Liu, Xin and Xu, Jun and Yue, Huanjing and Yang, Jingyu},
  journal={IEEE Transactions on Image Processing},
  year={2025},
  publisher={IEEE}
}

@article{mao2024generative,
  title={Generative Transformer for Accurate and Reliable Salient Object Detection},
  author={Mao, Yuxin and Zhang, Jing and Wan, Zhexiong and Tian, Xinyu and Li, Aixuan and Lv, Yunqiu and Dai, Yuchao},
  journal={IEEE Transactions on Circuits and Systems for Video Technology},
  volume={35},
  number={2},
  pages={1041--1054},
  year={2025},
  doi={10.1109/TCSVT.2024.3469286},
  publisher={IEEE}
}

@article{wang2024alignment,
  title={Alignment-free RGBT salient object detection: Semantics-guided asymmetric correlation network and a unified benchmark},
  author={Wang, Kunpeng and Lin, Danying and Li, Chenglong and Tu, Zhengzheng and Luo, Bin},
  journal={IEEE Transactions on Multimedia},
  volume={26},
  pages={10692--10707},
  year={2024},
  publisher={IEEE}
}

@article{zhu2024separate,
  title={Separate first, then segment: An integrity segmentation network for salient object detection},
  author={Zhu, Ge and Li, Jinbao and Guo, Yahong},
  journal={Pattern Recognition},
  volume={150},
  pages={110328},
  year={2024},
  publisher={Elsevier}
}

@inproceedings{feng2025residual,
  title={Residual Diffusion Deblurring Model for Single Image Defocus Deblurring},
  author={Feng, Haoxuan and Zhou, Haohui and Ye, Tian and Chen, Sixiang and Zhu, Lei},
  booktitle={Proceedings of the AAAI Conference on Artificial Intelligence},
  volume={39},
  number={3},
  pages={2960--2968},
  year={2025}
}

@article{chen2023hierarchical,
  title={Hierarchical integration diffusion model for realistic image deblurring},
  author={Chen, Zheng and Zhang, Yulun and Liu, Ding and Gu, Jinjin and Kong, Linghe and Yuan, Xin and others},
  journal={Advances in Neural Information Processing Systems},
  volume={36},
  pages={29114--29125},
  year={2023}
}

@inproceedings{peebles2023scalable,
  title={Scalable Diffusion Models with Transformers},
  author={Peebles, William and Xie, Saining},
  booktitle={Proceedings of the IEEE/CVF International Conference on Computer Vision},
  pages={4195--4205},
  year={2023}
}

@inproceedings{chang2025waterdiffusion,
  title={WaterDiffusion: Learning a Prior-involved Unrolling Diffusion for Joint Underwater Saliency Detection and Visual Restoration},
  author={Chang, Laibin and Wang, Yunke and Deng, Longxiang and Du, Bo and Xu, Chang},
  booktitle={Proceedings of the AAAI Conference on Artificial Intelligence},
  volume={39},
  number={2},
  pages={1998--2006},
  year={2025}
}

@inproceedings{li2025fscdiff,
  title     = {{FSCDiff}: Frequency-Spatial Entangled Conditional Diffusion Model for Underwater Salient Object Detection},
  author    = {Li, Hua and Lin, Gaowei and Li, Zhiyuan and Kwong, Sam and Cong, Runmin},
  booktitle = {Proceedings of the 33rd ACM International Conference on Multimedia},
  pages     = {8379--8388},
  year      = {2025},
  doi       = {10.1145/3746027.3755467}
}

@inproceedings{li2026waterflow,
  title     = {{WaterFlow}: Explicit Physics-Prior Rectified Flow for Underwater Saliency Mask Generation},
  author    = {Li, Runting and Lian, Shijie and Li, Hua and Li, Yutong and Wu, Wenhui and Kwong, Sam},
  booktitle = {ICASSP},
  year      = {2026}
}

@inproceedings{zhao2020suppress,
  title={Suppress and Balance: A Simple Gated Network for Salient Object Detection},
  author={Zhao, Xiaoqi and Pang, Youwei and Zhang, Lihe and Lu, Huchuan and Zhang, Lei},
  booktitle={Proceedings of the European Conference on Computer Vision},
  pages={35--51},
  year={2020},
  organization={Springer}
}

@inproceedings{pang2020multi,
  title={Multi-Scale Interactive Network for Salient Object Detection},
  author={Pang, Youwei and Zhao, Xiaoqi and Zhang, Lihe and Lu, Huchuan},
  booktitle={Proceedings of the IEEE/CVF Conference on Computer Vision and Pattern Recognition},
  pages={9413--9422},
  year={2020}
}

@inproceedings{wei2020label,
  title={Label Decoupling Framework for Salient Object Detection},
  author={Wei, Jun and Wang, Shuhui and Wu, Zhe and Su, Chi and Huang, Qingming and Tian, Qi},
  booktitle={Proceedings of the IEEE/CVF Conference on Computer Vision and Pattern Recognition},
  pages={13025--13034},
  year={2020}
}

@article{wu2022edn,
  title={{EDN}: Salient Object Detection via Extremely-Downsampled Network},
  author={Wu, Yu-Huan and Liu, Yun and Zhang, Le and Cheng, Ming-Ming and Ren, Bo},
  journal={IEEE Transactions on Image Processing},
  volume={31},
  pages={3125--3136},
  year={2022},
  publisher={IEEE}
}

@inproceedings{lee2022tracer,
  title={{TRACER}: Extreme Attention Guided Salient Object Tracing Network},
  author={Lee, Min Seok and Shin, Woojin and Han, Sung Won},
  booktitle={Proceedings of the AAAI Conference on Artificial Intelligence},
  volume={36},
  pages={12993--12994},
  year={2022}
}

@article{zhuge2022salient,
  title={Salient Object Detection via Integrity Learning},
  author={Zhuge, Mingchen and Fan, Deng-Ping and Liu, Nian and Zhang, Dingwen and Xu, Dong and Shao, Ling},
  journal={IEEE Transactions on Pattern Analysis and Machine Intelligence},
  volume={45},
  number={3},
  pages={3738--3752},
  year={2022},
  publisher={IEEE}
}

@ARTICLE{yun2022selfreformer,
  author={Yun, Yi Ke and Lin, Weisi},
  journal={IEEE Transactions on Multimedia}, 
  title={Towards a Complete and Detail-Preserved Salient Object Detection}, 
  year={2024},
  volume={26},
  number={},
  pages={4667-4680},
  doi={10.1109/TMM.2023.3325731}
  }

@article{ma2023boosting,
  title={Boosting Broader Receptive Fields for Salient Object Detection},
  author={Ma, Mingcan and Xia, Changqun and Xie, Chenxi and Chen, Xiaowu and Li, Jia},
  journal={IEEE Transactions on Image Processing},
  volume={32},
  pages={1026--1038},
  year={2023},
  publisher={IEEE}
}

@inproceedings{li2024size,
  title={Size-Invariance Matters: Rethinking Metrics and Losses for Imbalanced Multi-Object Salient Object Detection},
  author={Li, Feng and others},
  booktitle={Proceedings of the International Conference on Machine Learning},
  pages={28989--29021},
  year={2024}
}

@inproceedings{gao2024multi,
  title={Multi-Scale and Detail-Enhanced Segment Anything Model for Salient Object Detection},
  author={Gao, Shixuan and Zhang, Pingping and Yan, Tianyu and Lu, Huchuan},
  booktitle={Proceedings of the 32nd ACM International Conference on Multimedia},
  pages={9894--9903},
  year={2024},
  doi={10.1145/3664647.3680650}
}

@article{cong2026breaking,
  title={Breaking Barriers, Localizing Saliency: A Large-Scale Benchmark and Baseline for Condition-Constrained Salient Object Detection},
  author={Cong, Runmin and Chen, Zhiyang and Fang, Hao and Kwong, Sam and Zhang, Wei},
  journal={IEEE Transactions on Pattern Analysis and Machine Intelligence},
  volume={48},
  number={4},
  pages={4167--4183},
  year={2026},
  doi={10.1109/TPAMI.2025.3642893},
  publisher={IEEE}
}

@inproceedings{luo2024vscode,
  title={{VSCode}: General Visual Salient and Camouflaged Object Detection with 2D Prompt Learning},
  author={Luo, Ziyang and others},
  booktitle={Proceedings of the IEEE/CVF Conference on Computer Vision and Pattern Recognition},
  pages={17169--17180},
  year={2024}
}

@inproceedings{liu2022modeling,
  title={Modeling Aleatoric Uncertainty for Camouflaged Object Detection},
  author={Liu, Jiawei and Zhang, Jing and Barnes, Nick},
  booktitle={Proceedings of the IEEE/CVF Winter Conference on Applications of Computer Vision},
  pages={1445--1454},
  year={2022}
}

@inproceedings{pang2022zoom,
  title={Zoom In and Out: A Mixed-Scale Triplet Network for Camouflaged Object Detection},
  author={Pang, Youwei and Zhao, Xiaoqi and Xiang, Tian-Zhu and Zhang, Lihe and Lu, Huchuan},
  booktitle={Proceedings of the IEEE/CVF Conference on Computer Vision and Pattern Recognition},
  pages={2160--2170},
  year={2022}
}

@article{fu2023masnet,
  title={{MASNet}: A Robust Deep Marine Animal Segmentation Network},
  author={Fu, Zhenqi and Chen, Runmin and Huang, Yue and Cheng, Enhui and Ding, Xinghao and Ma, Kin-Keung},
  journal={IEEE Journal of Oceanic Engineering},
  volume={49},
  number={3},
  pages={1104--1115},
  year={2023},
  publisher={IEEE}
}

@inproceedings{zheng2021rethinking,
  title={Rethinking Semantic Segmentation from a Sequence-to-Sequence Perspective with Transformers},
  author={Zheng, Sixiao and Lu, Jiachen and Zhao, Hengshuang and Zhu, Xiatian and Luo, Zekun and Wang, Yabiao and Fu, Yanwei and Feng, Jianfeng and Xiang, Tao and Torr, Philip H. S. and Zhang, Li},
  booktitle={Proceedings of the IEEE/CVF Conference on Computer Vision and Pattern Recognition},
  pages={6881--6890},
  year={2021}
}

@article{he2023h2former,
  title={{H2Former}: An Efficient Hierarchical Hybrid Transformer for Medical Image Segmentation},
  author={He, Ao and Wang, Kai and Li, Tao and Du, Chengkun and Xia, Sihua and Fu, Huazhu},
  journal={IEEE Transactions on Medical Imaging},
  volume={42},
  number={9},
  pages={2763--2775},
  year={2023},
  publisher={IEEE}
}

@inproceedings{kirillov2023segment,
  title={Segment Anything},
  author={Kirillov, Alexander and Mintun, Eric and Ravi, Nikhila and Mao, Hanzi and Rolland, Chloe and Gustafson, Laura and Xiao, Tete and Whitehead, Spencer and Berg, Alexander C. and Lo, Wan-Yen and Doll{\'a}r, Piotr and Girshick, Ross},
  booktitle={Proceedings of the IEEE/CVF International Conference on Computer Vision},
  pages={4015--4026},
  year={2023}
}

@inproceedings{chen2023samadapter,
  title={{SAM-Adapter}: Adapting Segment Anything in Underperformed Scenes},
  author={Chen, Tianrun and Zhu, Lanyun and Deng, Chaotao and Cao, Runlong and Wang, Yan and Zhang, Shangzhan and Li, Zejian and Sun, Lingyun and Zang, Ying and Mao, Papa},
  booktitle={Proceedings of the IEEE/CVF International Conference on Computer Vision Workshops},
  pages={3367--3375},
  year={2023},
  doi={10.1109/ICCVW60793.2023.00361}
}

@inproceedings{yan2024massam,
  title={MAS-SAM: segment any marine animal with aggregated features},
  author={Yan, Tianyu and Wan, Zifu and Deng, Xinhao and Zhang, Pingping and Liu, Yang and Lu, Huchuan},
  booktitle={Proceedings of the Thirty-Third International Joint Conference on Artificial Intelligence},
  pages={6886--6894},
  year={2024}
}

@article{zhou2026turbidity,
  title={Turbidity-Similarity Decoupling: Feature-Consistent Mutual Learning for Underwater Salient Object Detection},
  author={Zhou, Wujie and Tang, Beibei and Cong, Runmin and Jiang, Qiuping},
  journal={IEEE Transactions on Image Processing},
  volume={35},
  pages={495--510},
  year={2026},
  doi={10.1109/TIP.2025.3648880}
}

@inproceedings{li2021mas3k,
  author    = {Lin Li and Eric Rigall and Junyu Dong and Geng Chen},
  title     = {{MAS3K}: An Open Dataset for Marine Animal Segmentation},
  booktitle = {Benchmarking, Measuring, and Optimizing},
  series    = {Lecture Notes in Computer Science},
  volume    = {12614},
  pages     = {194--212},
  publisher = {Springer},
  address   = {Cham},
  year      = {2021},
  doi       = {10.1007/978-3-030-71058-3_12}
}

@article{liu2020realworld,
  title={Real-World Underwater Enhancement: Challenges, Benchmarks, and Solutions Under Natural Light},
  author={Liu, Risheng and Fan, Xin and Zhu, Ming and Hou, Minjun and Luo, Zhongxuan},
  journal={IEEE Transactions on Circuits and Systems for Video Technology},
  volume={30},
  number={12},
  pages={4861--4875},
  year={2020},
  doi={10.1109/TCSVT.2019.2963772},
  publisher={IEEE}
}

@article{fu2023learning,
  title={Learning Heavily-Degraded Prior for Underwater Object Detection},
  author={Fu, Chenping and Fan, Xin and Xiao, Jiewen and Yuan, Wanqi and Liu, Risheng and Luo, Zhongxuan},
  journal={IEEE Transactions on Circuits and Systems for Video Technology},
  volume={33},
  number={11},
  pages={6887--6896},
  year={2023},
  doi={10.1109/TCSVT.2023.3271644},
  publisher={IEEE}
}

@article{lu2024speedup,
  title={Speed-Up {DDPM} for Real-Time Underwater Image Enhancement},
  author={Lu, Siqi and Guan, Fengxu and Zhang, Hanyu and Lai, Haitao},
  journal={IEEE Transactions on Circuits and Systems for Video Technology},
  volume={34},
  number={5},
  pages={3576--3588},
  year={2024},
  doi={10.1109/TCSVT.2023.3314767},
  publisher={IEEE}
}

@article{jia2026vit,
  title={ViT-UWA: Vision Transformer Underwater-Adapter for Dense Predictions Beneath the Water Surface},
  author={Jia, Yuheng and Lin, Qirui and Li, Hua and Li, Yutong and Kwong, Sam and Cong, Runmin},
  journal={IEEE Transactions on Image Processing},
  year={2026},
  publisher={IEEE}
}

\end{document}